\documentclass{JASSS}
	
\title{Neetyabhas: A Framework for Uncertainty-Aware Public Policy Optimization in Rational Agent-Based Models}

\reviewcopy{false}

\author{Janani Venugopalan}
\author{Gaurav Deshkar}
\author{Rishabh Gaur}
\author{Harshal Hayatnagarkar}
\author{Jayanta Kshirsagar}

\affil{Engineering for Research (e4r), Thoughtworks Technologies, Binarius Building, Deepak Complex, Shastrinagar, Yerawada, Pune, Maharashtra, India - 411006}

\email{janani.venugopalan@thoughtworks.com}

\usepackage{natbib}
	\setcitestyle{authoryear,round,aysep={}}
	
\usepackage{graphicx}%
\usepackage{multirow}%
\usepackage{amsmath,amssymb,amsfonts}%
\usepackage{amsthm}%
\usepackage{mathrsfs}%
\usepackage[title]{appendix}%
\usepackage{xcolor}%
\usepackage{textcomp}%
\usepackage{manyfoot}%
\usepackage{booktabs}%
\usepackage{algorithm}%
\usepackage{algorithmicx}%
\usepackage{algpseudocode}%
\usepackage{listings}%
\usepackage{graphicx}
\usepackage{subcaption}
\usepackage{multicol}
\usepackage{xr}
\usepackage{cleveref}
\usepackage{subfiles}
\usepackage{bm}

\let\oldcite\cite
\renewcommand{\cite}[1]{%
  \PackageWarning{natbib}{The command \texttt{\string\cite} is deprecated. Use \texttt{\string\citep} instead.}%
  \oldcite{#1}%
}

\begin{document}
\maketitle 



\begin{abstract}
\textbf{Purpose:}
In response to the COVID-19 pandemic, the World Health Organization (WHO) has endorsed an 18-point non-pharmaceutical intervention strategy, encompassing diverse measures like lockdowns, rapid vaccinations, and public transportation closures. While these interventions have demonstrated effectiveness in curtailing virus transmission and shaping public health policies, they have also introduced significant adverse consequences, notably the economic strain imposed by lockdown measures. Existing research often overlooks individual preferences and behaviors when designing balanced multifaceted policy frameworks. Moreover, prevalent studies tend to assume precise infection measurements and flawless policy execution, disregarding the substantial role of uncertainties and errors.

\textbf{Methods:}
To address these challenges, we present an integrative approach that incorporates uncertainties in both the measurement of the epidemic's state (infections and hospitalizations) and the implementation of policy directives. As proof of concept, we construct a simulation model involving a small community of 1000 individuals. These individuals make real-time decisions concerning mask-wearing, vaccination, and shopping behaviors as the epidemic unfolds. Policymakers can deploy interventions, such as lockdowns, immunization controls, and mask mandates, guided by observations of infections, hospitalizations, and economic indicators. This approach leverages hierarchical reinforcement learning agents, featuring deep Q-networks while extending uncertainty-aware deep deterministic policy gradient variants (DDPG and TD3) to inform individual decisions and policymaker choices.

\textbf{Results:}
Our simulation models effectively demonstrate the capacity to manage the epidemic's progression. Notably, the adoption of masking and vaccinations emerges as particularly promising strategies, effectively mitigating both the peak height and the duration of the outbreak. By accounting for individual behaviors, policy uncertainty, and multifaceted interventions, our approach proves successful in dynamically controlling the epidemic's impact.

\textbf{Conclusions:}
In conclusion, our integrative approach addresses the limitations of prevailing research by considering uncertainties and individual behaviors within multifaceted policy frameworks. Through simulation experiments, we illustrate the efficacy of our approach in controlling the COVID-19 epidemic. The adoption of masking and vaccinations emerges as pivotal measures in reducing the outbreak's severity. This research underscores the importance of accounting for uncertainties and individual decisions in shaping effective public health interventions in the face of complex pandemics.

\end{abstract}


\parano{}


\section{Introduction}
The COVID-19 pandemic has had devastating consequences worldwide, with over 460 million infections and 6 million deaths recorded by October 2022 ~\cite{wu2022economic}. Additionally, the pandemic caused a significant 3\% decline in global GDP in the year 2020 alone \cite{wu2022economic}. To combat the spread of the infection and mitigate its impact, the World Health Organization (WHO) has devised an 18-point non-pharmaceutical intervention (NPI) scheme \cite{yang2022computational}, encompassing measures such as lockdowns, rapid vaccination programs, public-transport closures, and economic stimulus \cite{perra2021non}. 

While these interventions have proven beneficial in controlling the outbreak, some, such as prolonged lockdowns and public transport closures, have also resulted in adverse effects, such as exacerbated economic disparities and compromised health outcomes due to overcrowding  \cite{bavli2020harms}. Governments worldwide face the challenge of determining the optimal level of intervention, given the numerous location-specific and culture-specific variables and factors involved.

To address this complex decision-making process, simulations have emerged as valuable tools, aiding governments and policymakers in making informed choices using these simulations \cite{venkatramanan2018using}. However, optimizing interventions to control an epidemic requires the consideration of multiple factors. Recent artificial intelligence and machine learning research \cite{payedimarri2021prediction, brodeur2021literature, yang2022computational, weltz2022reinforcement} on epidemic optimization through simulations has been abundant  \cite{puron2016opportunities, 10.1007/978-3-319-64322-9_9, wan2022advances, yang2022computational}, employing techniques such such as dynamic programming \cite{weltz2022reinforcement}, v-learning \cite{luckett2019estimating}, q-learning (deep and table based) \cite{khadilkar2020optimising, kwak2021deep, song2020reinforced, khadilkar2020optimising, ohi2020exploring, colas2021epidemioptim, bampa2022learning}, and greedy algorithms \cite{minutoli2020preempt}. These techniques scale poorly when the number of choices for optimization is large.

Reinforcement learning techniques, such as Deep Deterministic Policy Gradient (DDPG) \cite{chadi2022reinforcement}, Twin Delayed DDPG (TD3) \cite{chadi2022reinforcement}, and Proximal Policy Optimization (PPO) \cite{chadi2022reinforcement}, have been utilized to handle the challenges of large-scale optimization. Nevertheless, these approaches often neglect uncertainties in recommended actions and measured epidemic spread. Moreover, they do not account for the confounding effects of personal decisions made by rational agents when formulating policy decisions. Additionally, there is a lack of formal measures to compare and contrast different models in terms of reliability or trust.

This study introduces an epidemic model that incorporates two levels of decision-making: individual and public policy. To address personal decision-making, a Deep-Q-Network (DQN) based decision maker is utilized. For the public-policy level, a novel uncertainty-aware continuous reinforcement learning model, comprising uncertainty-aware DDPG and Twin Delayed DDPG, is proposed. To evaluate and compare these algorithms, metrics of reliability \cite{rl_reliability_metrics} and trust \cite{Sequeira2020} for epidemic optimization are extended and employed. In the interest of promoting further research in this area, the authors plan to release the code, simulation model, optimization methodology, and hyper-parameters utilized in the study as open-source resources. This will facilitate scalability, consideration of multiple objectives, and the enhancement of simulation models in future endeavors.

\section{Related Work}
The use of modeling and simulations for the discovery of optimal policies has grown significantly over the years as the models have become more realistic and their adoption rates have grown. Agent-based models (ABMs) are effective in capturing the population heterogeneity, relationships and stochasticity inherent in real-world interactions among individuals \cite{deangelis2019decision}. Researchers have employed ABMs to simulate disease transmission dynamics \cite{kerr2021covasim}, accounting for various factors such as age-stratified populations, multi-vaccine scenarios, and behavioral responses like mask-wearing and shopping habits \cite{gnanvi2021reliability}. These models have the advantage of being adaptable to specific contexts, enabling the study of localized outbreaks, and tailored interventions, and policy discovery. However, a significant challenge in the widespread adoption of automated policy discovery tools for ABMs is the presence of multiple conflicting optimization criteria and a large number of locale-specific factors that need to be optimized.

Several approaches have been proposed for suggested for public policy discovery, each with its strengths and limitations. Several Q learning based reinforcement learning techniques have been used for policy discovery in situations where the actions are discrete \cite{payedimarri2021prediction, brodeur2021literature, yang2022computational, weltz2022reinforcement, luckett2019estimating, khadilkar2020optimising, kwak2021deep, song2020reinforced, khadilkar2020optimising, ohi2020exploring, colas2021epidemioptim, bampa2022learning}. These studies are very useful where the number of interventions are limited e.g. type of masks worn, yes/no decision on shopping, decision for going out, etc. However, this forms a major limitation in a realistic public health scenario, where there are continuous factors such as the number of vaccinations needed and the duration of lockdowns \cite{lillicrap2015continuous}. To move towards the continuous space, some of the Q learning-based work has been extended using techniques such as restless bandits \cite{mate2021risk}, soft-Actor Critic \cite{kompella2020reinforcement}, collaborative RL \cite{bednarski2021collaborative}, recurrent neural networks based RL \cite{ge2020recurrent}, and contextual bandits \cite{awasthi2022vacsim}.  While the limitation of the discrete actions spaces has been mitigated with these studies, there are concerns regarding limited simulation objectives (e.g., vaccination, lockdown, economy), scale (few thousand individuals), or use of equation-based models unsuitable for intervention studies. Some of the recent contributions include continuous reinforcement learning frameworks such as Deep Deterministic Policy Gradient (DDPG), Twin Delayed DDPG, and Proximal Policy Optimization (PPO) \cite{chadi2022reinforcement, schulman2017proximal} for continuous action spaces. While these studies address the discrete and continuous action spaces, they do not include hierarchical decision-making. For example, while policymakers make public policy to mitigate the far-reaching consequences of the pandemic, individuals make local decisions to best cope with the pandemic and economic stress while staying compliant with public policy. Studies by Dong \emph{et al.} \cite{dong2020hierarchical} simulate individual and household decisions using a hierarchical reinforcement learning framework, that incorporates different levels of local choices. However, they do not account for the public health policymaker's role in decision-making. Zheng \emph{et. al.} develop a multi-level model for economics where the control is again at a local and regional level. A common drawback across all these studies is that they assume that all the observations made and decisions implemented are accurate and no uncertainty is accounted for.

Uncertainty is inherent in policymaking due to unpredictable events, data limitations, and variations in human behavior. Aleatory uncertainty pertains to the inherent randomness in epidemic processes and input parameters, while epistemic uncertainty arises from state measurement, and model structure and parameter estimation. Traditional policy optimization approaches typically neglect these uncertainties, leading to suboptimal decisions or potential policy failures. Recent research has begun to address these issues, incorporating techniques to handle uncertainty, such as stochastic optimization and robust decision-making. However, there remains a gap in extending these approaches to account for both aleatory and epistemic uncertainty concurrently, which could enhance the reliability and adaptability of policy recommendations. Very few studies in public health discuss uncertainty in decision making and of those, nearly 70\% are imaging studies and use Monte Carlo dropouts to mitigate and quantify uncertainty \cite{loftus2022uncertainty}. Other techniques such as distribution-based Q learning \cite{browning2021uncertainty}, determining the distributions of actions and states \cite{nanayakkara2022unifying}, uncertainty decomposition for uncertainty quantification \cite{festor2021enabling}, and variational long-short-term memory (LSTM) based contact networks to approximate the spread of epidemics under incomplete information \cite{feng2022precise} have been proposed for medical decision making under uncertainty. However, most of these techniques are used where the uncertainty distributions are known or for uncertainty quantification. A recent study by Tessler \emph{et al.} \cite{tessler2019action} developed fundamental theory and RL techniques that can work with uncertain actions. This study does not demonstrate any applications and only accounts for aleatory uncertainty (uncertainty caused due to measurement errors) but not epistemic uncertainty (uncertainty in state measurement and model weights due to less and uncertain data). While aleatory uncertainty, as handled by Tessler \emph{et al.}, can account for uncertain actions, epistemic uncertainty can account for uncertainty in both state and action. In our study, we adopt and extend the techniques demonstrated by Tessler \emph{et. al.} for both types of uncertainty as seen in the epidemiology domain.

After the techniques are established, measuring the efficacy and trustworthiness of policy decisions is crucial to ensure effective implementation and public acceptance. Existing literature has proposed various metrics, including reliability metrics \cite{rl_reliability_metrics} and trust \cite{Sequeira2020}, to assess the performance and credibility of reinforcement learning models. Reliability metrics quantify the consistency and accuracy of model predictions, while trust metrics gauge the confidence in model outputs based on historical performance and validation. Incorporating such metrics in policy optimization frameworks can offer policymakers insights into the trade-offs between economic considerations and public health outcomes. As a result, the present study presents a significant advancement in agent-based epidemiological modeling by devising a comprehensive model that integrates individual and policy-level decision-making processes. By simulating various scenarios, such as lockdowns, age-stratified multi-vaccination strategies, economic dynamics, and behavioral variations, the model provides a holistic perspective on disease transmission dynamics in small communities. Moreover, we introduce novel methods for optimizing public policies under uncertainty, encompassing both aleatory and epistemic uncertainties. This extension of existing techniques enhances the robustness and practicality of policy recommendations, thus bridging an important gap in current research. Additionally, the study explores the application and extension of reliability and trust metrics for evaluating different algorithms in epidemic optimization, contributing valuable insights into policy evaluation and selection. 

\section{Key Contributions}
The key contributions of this study are as follows:
\begin{itemize}
    \item \textbf{Agent-Based Epidemic Model}: We have devised an agent-based epidemic model that captures individual and policy-level decision-making dynamics in a small community. The model incorporates essential elements such heterogeneous agents, their daily schedules, interactions, and travel. In addition, we have a basic economic model, and variations in masking and shopping behavior. Our model also supports interventions such as lockdowns, and age-stratified multiple vaccination programs. The model in this study is relatively minimalist, mainly to demonstrate rational agent behavior and the policy optimization around it. We intend to scale this model for larger communities, cities, and even combinations of urban and rural areas in future expansions. The BharatSim framework \cite{bharatsim} used in implementing this model is flexible and performant enough to accommodate highly complex and diverse simulations at scale.
    
    \item \textbf{Novel Methods for Public Policy Optimization:} We propose innovative approaches to optimize public policy decisions that strike a balance between economic considerations and preserving human lives. Our methods explicitly account for both aleatory uncertainty (related to uncertain inputs) and epistemic uncertainty (arising from uncertain models and weightings). This extension of existing techniques enhances the robustness and reliability of policy recommendations, especially in the face of unpredictable and dynamic situations.
    \item \textbf{Enhanced Metrics for Epidemic Optimization:} In comparing and contrasting the different algorithms presented in this study, we employ and extend reliability metrics \cite{rl_reliability_metrics} and trust \cite{Sequeira2020} assessment techniques for the field of epidemic optimization. These metrics enable a comprehensive evaluation of the efficacy and trustworthiness of various policy choices, providing policymakers with a more nuanced understanding of their implications and performance. 
\end{itemize}

We organize the remainder of the paper as follows: We introduce the formalized problem statement in Section \ref{formal_problem}; It is followed by the proposed model and solution to address the problem statement (Section \ref{model}); Then we discuss the evaluation criteria and the results of the evaluation (Section \ref{evaluation}); Then we discuss the experiments performed in the epidemiology domain, their results and insights (Section \ref{experiments}); Finally we conclude the paper in Section \ref{conclusion}.

\section{Formalized Problem Statement} \label{formal_problem}
For the epidemiological domain, we frame the problem statement as a multi-objective optimization problem with uncertainty and free-acting agents. The details of those are as follows:

\subsection{Multi-objective Optimization Problem with Uncertainty and Free-Acting Agents}
Consider a small community $C$ comprising $N$ individuals, each denoted by $i \in \{1, 2, \ldots, N\}$. The goal is to devise a multifaceted intervention strategy $I = \{I_1, I_2, \ldots, I_{18}\}$ to combat an epidemic, aiming to minimize both the peak height of infections and the economic impact. This optimization problem accounts for individual preferences and behaviors, as well as uncertainties associated with measurements and policy implementation.

\subsection{Decision Variables}
For each individual $i$ and each time step $t$, let $x_{i,t}$ represent a decision vector with components representing:
\begin{itemize}
    \item Mask-wearing behavior ($x_{i,t}^{\text{mask}}$) as a categorical variable (1, 2 for wearing a particular type of mask, 0 for not wearing).
    \item Vaccination status ($x_{i,t}^{\text{vaccine}}$) as a categorical variable (1, 2 for wearing a particular type of vaccine, 0 for not vaccinated).
    \item Shopping frequency ($x_{i,t}^{\text{shop}}$) as a binary variable representing the need to shop (1 for shopping and 0 for not shopping).
\end{itemize}

\subsection{Objective Functions}
\begin{itemize}
\item We minimize peak and total infections. Given $I_{i,t}^{\text{inf}}$ as the number of infections at time $t$ for individual $i$, the peak and total infections are governed by the masking ($x_{i,t}^{\text{mask}}$), vaccination ($x_{i,t}^{\text{vaccine}}$), and shopping behaviors ($x_{i,t}^{\text{shop}}$). 
\item We minimize the economic impact by reducing the blow poverty line individuals. It is governed by shopping behaviors ($x_{i,t}^{\text{shop}}$)
\end{itemize}

\subsection{Constraints}
\begin{itemize}
\item \textbf{Uncertainty in Measurements}
The measurements of the epidemic such as infections, hospitalizations, and the below poverty line individuals are uncertain and influenced by the true values $I_{i,t}^{\text{inf, true}}$, $H_{i,t}^{\text{true}}$, and $BPL_{i,t}^{\text{true}}$ as well as uncertainty factors $\epsilon_{i,t}^{\text{inf}}$, $\epsilon_{i,t}^{\text{hospital}}$ and $\epsilon_{i,t}^{\text{BPL}}$ :
\[
I_{i,t} = I_{i,t}^{\text{inf, true}} + \epsilon_{i,t}^{\text{inf}}, \quad H_{i,t} = H_{i,t}^{\text{true}} + \epsilon_{i,t}^{\text{hospital}}, \quad BPL_{i,t} = BPL_{i,t}^{\text{true}} + \epsilon_{i,t}^{\text{BPL}}
\]

\item \textbf{Uncertainty in Policy Implementation}
The effectiveness of policy interventions is uncertain and influenced by uncertainty factors $\epsilon_{i,t}^{\text{policy}}$:
\[
I_{i,t}^{\text{inf}} = f(I_{i,t-1}^{\text{inf}}, I_{i,t-1}^{\text{hospital}}, I_{t-1}^{\text{eco}}, I_{t-2}^{\text{eco}}, x_{i,t}^{\text{mask}}, x_{i,t}^{\text{vaccine}}, x_{i,t}^{\text{shop}}, \epsilon_{i,t}^{\text{policy}})
\]

\item \textbf{Individual Behavior Constraints}
Individual shopping frequency is constrained to a maximum value $S_{\text{max}}$:
\[
x_{i,t}^{\text{shop}} \leq S_{\text{max}}
\]
\end{itemize}

The challenges are further confounded by the fact that uncertainty bounds vary depending on a variety of factors and are not known. To address the above-mentioned challenge, our proposed model and solutions are detailed in the next sections.

\section{Proposed Model} \label{model}
\label{sec:epi-model}
\begin{figure}[ht]
  \centering
  \includegraphics[width=\columnwidth]{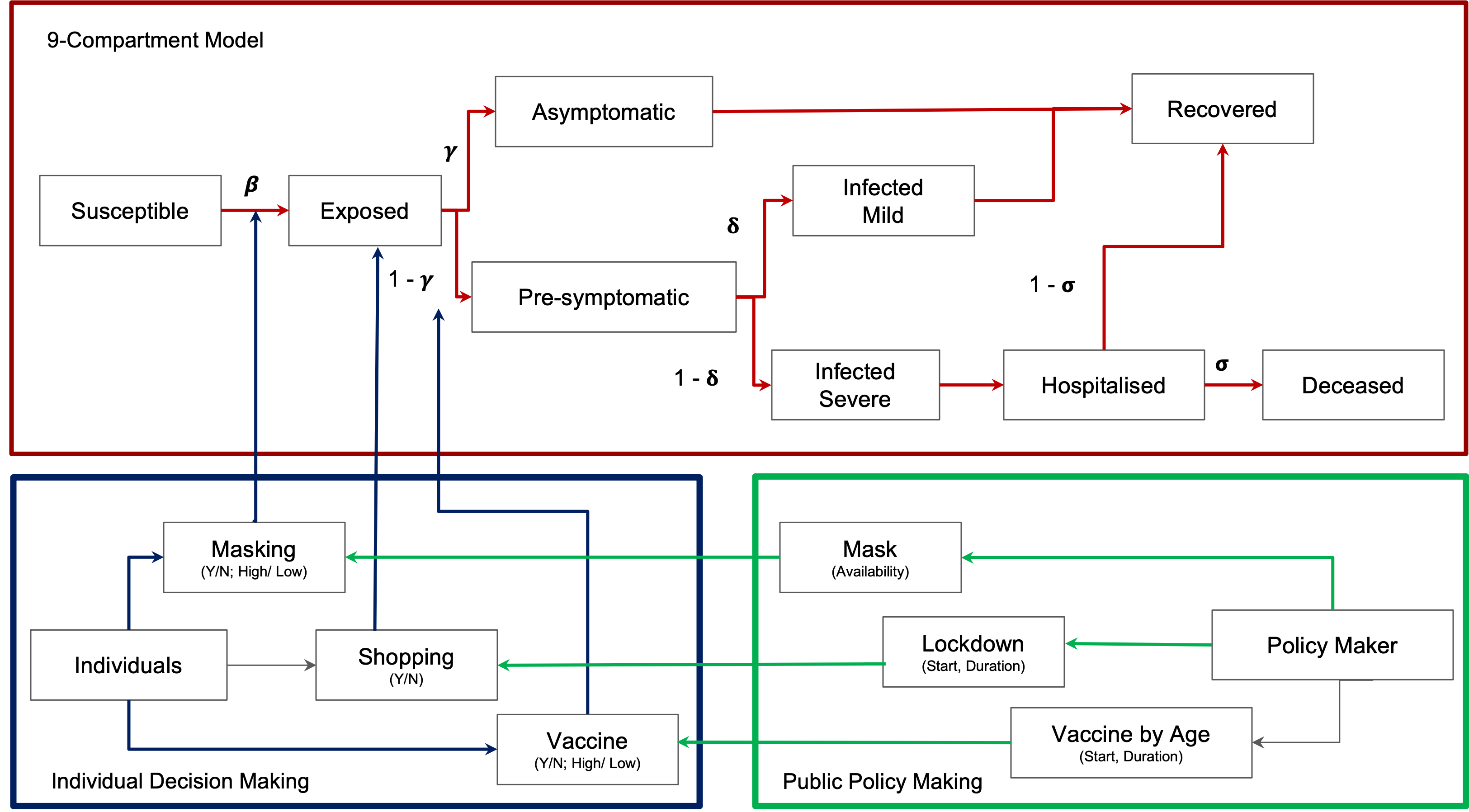}
  \caption{Agent-based Epidemic simulator with the 9-Compartment model, individual decision-making components, and public policy-making.}
  \label{fig:block_diagram}
\end{figure}

We propose a minimalist agent-based model ``(Figure \ref{fig:block_diagram})'' which simulates an epidemic in small communities with a thousand agents. It incorporates rational decision-making at the public health and individual levels. Our simulation has the following modules: 
\begin{itemize}
    \item \textbf{Individuals} to model people who make rational choices about their behavior as the epidemic unfolds. 
    \item \textbf{Geography} to model the co-locations of individuals based on a schedule.
    \item \textbf{Nine-compartment disease model} to model disease dynamics. \cite{tolles2020modeling} 
    \item \textbf{Intervention and policy} to model intervention and related approaches like lockdown, vaccination policy, etc. 
    \item \textbf{Economy/ House units} to demonstrate consumption and shopping behavior.
    \item \textbf{Individual decision-making units} using deep-Q-network (DQN) based learning agents.
    \item \textbf{Uncertainty-aware public policy decision-making units} using continuous state, action reinforcement learning agents. 
\end{itemize}

The details of each of these models is as follows:

\subsection{Individuals}
We model a small community with 1000 individuals as agents. Agents older than 30 are considered employed, and those younger than 30 are students using an open-source extensible simulation framework (Bharatsim \cite{bharatsim}). Every agent follows a schedule. A schedule is defined for 24 hours, with six vector ticks in the simulation. Agents spend 8 hours (2 vector ticks) at home, followed by 12 hours (3 vector ticks) at either the office or school based on their age, followed by 4 hours (1 vector tick) shopping or at home based on their preference. We chose these locations to demonstrate a minimalist model where people are moving to meet the same people routinely (offices, houses) and different people intermittently (shops).
    
\subsection{Geography}
Geography comprises five main areas: houses, offices, schools, shops, and hospitals. Based on the schedule and its attributes, an agent would be at home, office/school, or shop at any given time. Hospitalized individuals will spend all their time in the hospital.

\subsection{Nine-Compartment Epidemic Model}
The nine-compartment model handles the logic for disease transmission. We use a variation of the SEIR model \cite{hazra2021indsci} with 1000 agents, adapted for demonstrating the role of individual and public decision-making.  Every agent would be at exactly one and only one compartment at any point in time. The disease compartments are shown in ``Figure \ref{fig:block_diagram}'' \cite{hazra2021indsci}. The transition factors ($\beta, 1-\gamma, 1-\delta,  \sigma$) are age-stratified, as shown in ``Table.~\ref{tab_disease_age_stratified}'' \cite{kerr2021covasim, hazra2021indsci}. $\beta$, $\gamma$, $\delta$, and $\sigma$ is a transition factor to exposed, asymptomatic, infected-mild, and deceased compartments respectively as shown in ``Figure \ref{fig:block_diagram}''. The incubation period and the duration of the number of days spent in a particular compartment are derived from a log-normal distribution with the means and standard deviations shown in ``Table.~\ref{tab_duration_by_compartment}'' \cite{kerr2021covasim, childs2020impact,hazra2021indsci}.
    
    \begin{table}[htbp]
        \caption{Disease transition factor by age group}
        \begin{tabular}{|c|c|c|c|c|}
        \toprule
        \textbf{Age Group} & \textbf{\textit{ $\beta$ multiplier}}& \textbf{\textit{ 1 - $\gamma$ }}  & \textbf{\textit{ 1 - $\delta$ }}  & \textbf{\textit{ $\sigma$ }} \\
        \midrule
            0 - 9 & 0.34 & 0.5  & 0.0005  &  0.00002  \\
            10 - 19 & 0.67 & 0.55  & 0.00165  & 0.00002  \\
            20 - 29 & 1.0 & 0.6   & 0.00720 & 0.0001  \\
            30 - 39 & 1.0 & 0.65   & 0.02080 & 0.00032  \\    
            40 - 49 & 1.0 & 0.7   & 0.03430 &  0.00098  \\  
            50 - 59 & 1.0 & 0.75   & 0.07650 & 0.00265  \\ 
            60 - 69 & 1.0 & 0.8   & 0.13280 & 0.00766  \\  
            70 - 79 & 1.24 & 0.85   & 0.20655 & 0.02439  \\   
            80 - 89 & 1.47 & 0.9   & 0.24570 & 0.08292 \\  
            90 - 99 & 1.47 & 0.9   & 0.24570 & 0.16190  \\      
        \bottomrule
       
        \multicolumn{5}{l}{ \textbf{Note:} $\beta$ defined as 0.5 is multiplied by ``$\beta$ multiplier'' for age.}
        \end{tabular}
        \label{tab_disease_age_stratified}
    \end{table}

 \subsection{Interventions and Policy}
Specific interventions, such as lockdowns and vaccination, are used to prevent the spread of the epidemic. To impose and implement these interventions, public policies are defined. The model has the following interventions:
\begin{itemize}
    \item \textbf{Lockdown}: Most individuals stay home when the lockdown is applied. Only essential workers are allowed to work. There are also some lockdown violators. In our model, about 20\% of the population are essential workers, and about 10\% are lockdown violators. The related policy specifies the start day and end day of the lockdown.
    \item \textbf{Vaccination}: When an individual has been vaccinated, their chances of getting infected ($\beta$) are reduced based on the vaccine's effectiveness. A vaccinated individual's $\gamma$ is increased by 80\%. i.e., in case of infection, it is more likely for them to get an asymptomatic infection. The individual disease transmission factor is reduced by 20\%, which implies that they are less likely to pass on the disease. There are two types of vaccines in our model with tunable effectiveness. The population is age-stratified into three age groups 0-17, 18-59, and 60-99. The related policy specifies the number of available doses for each vaccine type and the start and end days of vaccination for each age group.
    \item \textbf{Masks}: When an individual wears a mask, their chances of getting infected ($\beta$) are reduced based on the mask's effectiveness. Our model has two types of masks. The related policy specifies the number of available masks for each mask type.
\end{itemize}

\begin{table}[htbp]
        \caption{Number of days spent in compartment}
        \begin{tabular}{|c|c|c|}
        \toprule
        \textbf{Compartment} & \textbf{\textit{ Mean}}& \textbf{\textit{ Standard Deviation  }} \\
        \midrule
            Exposed & 4.5  & 1.5  \\
            Asymptomatic & 8.0  & 2.0  \\    
            Pre-Symptomatic & 1.1 & 0.9  \\
            Infected Mild & 8.0  & 2.0  \\
            Infected Severe & 1.5 & 2.0 \\  
            Hospitalized & 18.1 & 6.3\\  
        \bottomrule
        \multicolumn{3}{l}{ \textbf{Note:} derived from log normal distribution.}
        \end{tabular}
        \label{tab_duration_by_compartment}
    \end{table}
 \subsection{Economy}
 We model a basic economic activity with the house as the central unit and essential items as the monetary unit. Each house has some stock of essential items (units), and each person consumes about six units daily. The average size of a housing unit is about four people. Initial stocks are distributed by a normal distribution with a mean of 130 units and a standard deviation of 20 units, which is about one week of stock for an average housing unit. When an agent goes shopping, it gets three days of stock for the average housing unit, i.e., 72 units of essential items added to the house's essential items stock to which the agent belongs. In case of lockdown or hospitalization agent can't go shopping. There are minimum and maximum limits set on the stock that every housing unit can hold. These limits are used to reward or penalize agents for excessive or inadequate shopping. The minimum limit is 1.25 per-day stock, i.e., 30 units, and the maximum is two weeks of stock, i.e., 360 units.


\subsection{Individual Decision Making} 
Our model's individuals (agents) can make critical decisions concerning their behavior as the epidemic unfolds. We model each agent in our simulation as Markov Decision Process (MDP) with discrete actions but continuous state space to account for decisions such as masking behavior (Y/N mask, type of mask), willingness for vaccination, and type preference (Y/N vaccination, type of vaccination) and shopping(Y/N). The state space includes infections, hospitalizations, etc.

The MDP here is defined as the 5-tuple ($S, A, P, R, \gamma$), where $s \in S$ is a finite state space, $a \in A$ is a discrete action space, $P \equiv P(s_{t+1}|s_{t}, a_{t})$ is a transition kernel. $R \equiv r(s_{t}, a_{t})$ is a reward function, and $\gamma \in (0, 1)$ is the discount factor. To account for continuous state and discrete action MDP,  we set up the MDP as a Deep-Q-Learning (DQN) based RL agent, where the rewards ($r(s_{t}, a_{t})$) are approximated using a $Q$ value given by Bellman Equation ``(Equation \ref{eq:q_regular})''. 

The DQN agent gets a set of observations and rewards/penalties from an environment in response to a set of actions/behaviors. An agent learns optimal behavior in an environment to obtain the maximum reward. We are using Deep Q-Networks (DQN) algorithm. It approximates a state-value function in Q-Learning with a neural network. In this work, we have used a variant of DQN to work with multiple agents \cite{tang2021microscopic}. In our model, the observations are the joint states of all the individuals from the epidemic model. The observations are as follows:

\textbf{Observation} :
\begin{itemize}
    \item{\textbf{Agent attributes/states}: age, infectedState, occupation, isEssentialWorker, violateLockdown, vaccineType, maskType, didShop, currentPlace, nextPlace, infectionCountInLocalNetwork, houseId, houseStock }  
    \item{\textbf{Global observation}:  currentTick, totalInfected, totalHospitalization, isLockdownOn.}
    \item {\textbf{Policy observation}: lockdownStart, lockdownEnd, maskAvailabilityPerType, vaccineAvailabilityPerType, vaccineStartStopByAgeGroup}
\end{itemize}

\textbf{Actions} : 
\begin{itemize}
    \item{\textbf{Mask preference}: No-mask, Type-1-mask, Type-2-mask }
    \item{\textbf{Vaccine preference}:  No-vaccine, Type-1-vaccine, Type-2-vaccine}
    \item {\textbf{Shopping preference}: No-shopping, Go-shopping}
\end{itemize}
 The action preferences for masks and vaccines are evaluated with the public policy and only allocated to the agent if the preferred mask and vaccine are available and the agent's age group is allowed to vaccinate. The shopping preference is executed based on the lockdown policy and the agent's hospitalization status. If the preferred action can't be performed for any reason mentioned above, the default for each action is No-mask, No-vaccine, and No-shopping.

\textbf{Rewards}: One of the primary rewards is when the agent transitions to an exposed state; it is a penalty. This penalty is calculated dynamically based on the percentage of currently infected agents. The penalty is ten times the infected agent percentage and a minimum of 30. Rest of the rewards are described in ``(Table \ref{tab0})'':
 \begin{table}
 \caption{Reward Structure} \label{tab0}
 \centering
 \begin{tabular}{|p{0.55\linewidth}|p{0.3\linewidth}|}
  \toprule
     Condition & Rewards/Penalty  \\
 \midrule
    Agent stays Susceptible  & reward of 10 \\
    Agent wears mask type 1 & penalty of 4 \\
    Agent wears mask type 2 & penalty of 2 \\
    Agent does inadequate shopping & penalty of 5 \\
    Agent does excessive shopping & penalty of 5 \\
    Agent has adequate essential stock  & reward of 1 \\
\bottomrule
 \end{tabular}
 \end{table}
 
 The mask penalty models the discomfort of wearing a mask and the monetary cost of a mask. The reward of wearing a mask and taking the vaccine is implicit in the effect of the mask and the vaccine, i.e., reducing the chances of exposure.

\subsection{Uncertainty Aware Public Policy Decision Making}
\label{sec:public_policy}
 Policymakers make decisions regarding the vaccination policy (whom to vaccinate and when), mask availability (import/ increase manufacturing), and lockdowns (restricting movement). In our model, we consider the policymaker and the optimization of public policy as a Markov Decision Process (MDP) with a continuous action space to account for factors such as lockdown start and end days, vaccination start and end days, and mask availability. We also consider that while the directives of public policy (actions), such as the exact lockdown start, and vaccination, cannot be followed, the measurement of states, such as the number of infected, is also not precise. 

A continuous space-action MDP is defined as ($S, A, P, R, \gamma $), where $s \in S$ is a finite state space, $a \in A$ is a finite action space, $P \equiv P(s_{t+1}|s_{t}, a_{t})$ is a transition kernel which is continuous in $a$. $R \equiv r(s_{t}, a_{t})$ is a reward function continuous in $a$, and $\gamma \in (0, 1)$ is the discount factor. While in a traditional MDP, the actions ($a \in A$) and the rewards ($r(s_{t}, a_{t})$) are considered certain. In our model, we account for uncertainty in actions by applying ``ActionRobustRL'' by Tessler \emph{et al.} \cite{tessler2019action}. We then extend their work to account for uncertainty in states and actions.

\subsubsection{Uncertainty in actions}
We consider action robustness as the response to a scenario where there are perturbations in actions. In this work, we consider Noisy Action Robust MDP \cite{tessler2019action} (NR-MDP), in which a low amount of adverse perturbation is added to the action. This simulates a situation where individual actions taken (by some individuals) are contrary to the policymaker's directive. Some examples include lockdown violators and anti-vaccine groups. The task of the policymakers then becomes that of finding the optimal policy ($\pi^{*}$) such that

\begin{eqnarray}\label{eq:policy_action}
\pi_{N, \alpha}^{*}(a,s) = E_{b \sim \pi(.|s), \bar{b} \sim \bar{\pi}(.|s)}[1_{a}=(1-\alpha)b + \alpha\bar{b}]
\end{eqnarray}

where $\pi$ is the policy in the absence of noise, and $\bar{\pi}$ is the adversarial policy. $b$ is the action sampled from $\pi$, and $\bar{b}$ is the action sampled from $\bar{\pi}$. $\alpha \in [0, 1]$ is the uncertainty value determining the percentage of adversarial actions. 

As mentioned in Tessler \emph{et al.} \citep{tessler2019action}, there are no theoretical assumptions of stationary or lower bounds on the $\pi^{*}$. In addition, due to the implementation of the MDP using neural networks (Deep Deterministic Policy Gradient (DDPG) and Twin Delayed DDPG (TD3)), no theoretical bounds exist. When neural networks are used to approximate an MDP, the rewards ($r(s_{t}, a_{t})$) are approximated using a $Q$ function/ a critic neural network. The policy (($\pi(.|s)$) is approximated using an actor-network. In uncertain action, MDP, the policy($\pi^{*}(.|s)$)  is also governed by an adversarial network. While Tessler \emph{et. al} limit the heuristics to Deep Deterministic Policy Gradient (DDPG) (Supplement: Algorithm \ref{alg:nrmdp_ddpg}), in our work, we extend it to Twin Delayed DDPG (TD3) (Algorithm \ref{alg:nrmdp_td3}). 

\noindent
\begin{minipage}[t]{\textwidth}
\begin{algorithm}[H]
\caption{Uncertain Action TD3 Algorithm and Update}\label{alg:nrmdp_td3}
\footnotesize
\begin{algorithmic}[1]

\State \textbf{Input: } Actor update steps ($N$), uncertainty value $\alpha$ and discount factor $\gamma$

\State \textbf{Initialize: } Randomly initialize critic network $Q_{1,2}(s,a;\phi)$, actor $f(s;\theta)$ and adversary $\bar{f}(s;\bar{\theta})$

\State Initialize target networks $Q'_{1,2}(s,a;\phi^{-})$, $f'(s;\theta^{-})$ and $\bar{f}'(s;\bar{\theta}^{-})$ with weights $\phi^{-}$, $\theta^{-}$ and $\bar{\theta}^{-}$

\State Initialize replay buffer Replay

\For{episode in $0...M$}
 \State Reset to initial state $s_{0}$
 \For{$t = 1, T$}
  \State Sample action $a_{t} \sim \pi^{*}$; $a_{t} = (1 - \alpha)b + \alpha \bar{b}$
  \State $\tilde{a}_{t} = a_{t} +$ exploration noise
  \State Execute $\tilde{a}_{t}$ and observe reward $r_{t}$ and new state $s_{t+1}$
  \State Store transition ($s_{t}, \tilde{a}_{t}, r_{t}, s_{t+1}$) in Replay
  \State Sample a random mini-batch of transitions from replay
  \State Set $ y_{i} = r + \gamma[min(Q'(s',(1 - \alpha)f'(s;\theta^{-}) + \alpha \bar{f}'(s;\bar{\theta}^{-}))]$
  \State Update critic with critic loss 
  \State $L = \frac{1}{N}\sum_{i}([y_{i} - Q_{1}(s_{i},a_{i})]^2 + [y_{i} - Q_{2}(s_{i},a_{i})]^2)$
  \If{episode = $N$}
    \State Update actor $\theta \gets \nabla_{\theta} Q_{1}(s,(1 -\alpha)f(s;\theta) + \alpha\bar{f}(s;\bar{\theta}))$
    \State Update adversary 
    \State$\bar{\theta} \gets \nabla_{\bar{\theta}} Q_1{1}(s,(1 -\alpha)f(s;\theta) + \alpha\bar{f}(s;\bar{\theta}))$
    \State Update the target networks
    \State $\phi^{-} \gets \tau\phi + (1-\tau)\phi^{-}$
    \State $\theta^{-} \gets \tau\theta + (1-\tau)\theta^{-}$
    \State $\bar{\theta}^{-} \gets \tau\bar{\theta} + (1-\tau)\bar{\theta}^{-}$
  \EndIf
 \EndFor
\EndFor

\end{algorithmic}
\end{algorithm}
\end{minipage}

\subsubsection{Uncertainty in states}
We consider states and action robustness as the response to a scenario with perturbations, such as measurement uncertainty, noise, or model uncertainty. Some examples include uncertain measurement of the number of infections, vaccinations, or poverty levels.
To account for this, we model the critic network as a Bayesian Neural Network. 

Bayesian networks have been proposed previously to provide formalism to uncertainty-aware RL \cite{DBLP:journals/corr/GhavamzadehMPT16}. However, there are challenges due to the need to select appropriate priors during training and extensive Monte Carlo sampling during training and inferences which have restricted the use of Bayesian Networks. In other disciplines, using Bayesian Neural Networks \cite{goan2020bayesian}, variational inference \cite{wu2018deterministic}, and frameworks such as pyro \cite{ritter2022tyxe} have mitigated these challenges. In our work, we extend the use of Bayesian Neural Networks with variational inference to the DDPG mentioned above and TD3 to account for uncertainty in states.

\noindent
\begin{minipage}[t]{\textwidth}
\begin{algorithm}[H]
\caption{Uncertain State DDPG Algorithm and Update}\label{alg:ddpg_bayes}
\footnotesize
\begin{algorithmic}[1]

\State \textbf{Input: } Actor update steps ($N$), uncertainty value $\alpha$, discount factor $\gamma$, samples $S$

\State \textbf{Initialize: } Randomly initialize critic network $Q(s,a;\phi) \sim N(0,1)$, and actor $f(s;\theta)$

\State Initialize target networks $Q'(s,a;\phi^{-})$, and $f'(s;\theta^{-})$ with weights $\phi^{-}$, and $\theta^{-}$

\State Initialize replay buffer Replay

\For{episode in $0...M$}
 \State Reset to initial state $s_{0}$
 \For{$t = 1, T$}
  \State Sample action $a_{t} \sim \pi$; $a_{t} = b$
  \State $\tilde{a}_{t} = a_{t} +$ exploration noise
  \State Execute $\tilde{a}_{t}$ and observe reward $r_{t}$ and new state $s_{t+1}$
  \State Store transition ($s_{t}, \tilde{a}_{t}, r_{t}, s_{t+1}$) in Replay
  \State Sample a random mini-batch of transitions from replay
  \State Set $ y_{i} = r + \gamma[Q'(s',f'(s;\theta^{-}))]$
  \State where we sample  weights $\phi^{-}$ $S$ times and compute mean
  \State Update critic with critic loss $L = \frac{1}{N}\sum_{i}[y_{i} - Q(s_{i},a_{i})]^2$
  \State where we sample  weights $\phi$ $S$ times and compute mean
  \If{episode = $N$}
    \State Update actor $\theta \gets \nabla_{\theta} Q(s,f(s;\theta))$
    \State Update the target networks
    \State $\phi^{-} \gets \tau\phi + (1-\tau)\phi^{-}$
    \State $\theta^{-} \gets \tau\theta + (1-\tau)\theta^{-}$
  \EndIf
 \EndFor
\EndFor

\end{algorithmic}
\end{algorithm}
\end{minipage}
\newline
\newline
The critic neural network ($Q$) is modeled as a Bayesian neural work, where the weights $\phi$ are sampled from a Normal distribution with a trainable mean $\mu$ and standard deviation $\sigma$. For each training and inference of the critic network, we sample a set of weights $\phi$ and then compute the Expected value of $Q$. This value is plugged into the Bellman Equation ``(Equation \ref{eq:q_regular})'' to get the approximation of reward ($r(s_{t}, a_{t})$). 
\begin{eqnarray}\label{eq:q_regular}
Q^{\pi}(a_{t},s_{t}) = E_{r_{t}, s_{t+1} \sim E}[r(s_{t},a_{t}) + \gamma E_{a_{t+1} \sim \pi}[Q^{\pi}(a_{t+1},s_{t+1})]]
\end{eqnarray}

The heuristics for training the Deep Deterministic Policy Gradient (DDPG)  and Twin Delayed DDPG (TD3)  in Algorithms (Algorithm \ref{alg:ddpg_bayes}) and (Algorithm \ref{alg:td3_bayes})

\noindent
\begin{minipage}[t]{\textwidth}
\begin{algorithm}[H]
\caption{Uncertain State TD3 Algorithm and Update}\label{alg:td3_bayes}
\footnotesize
\begin{algorithmic}[1]

\State \textbf{Input: } Actor update steps ($N$), uncertainty value $\alpha$, discount factor $\gamma$ and samples $S$

\State \textbf{Initialize: } Randomly initialize critic network $Q_{1,2}(s,a;\phi) \sim N(0, 1)$, actor, and $f(s;\theta)$.

\State Initialize target networks $Q'_{1,2}(s,a;\phi^{-})$, and $f'(s;\theta^{-})$ with weights $\phi^{-}$, and $\theta^{-}$

\State Initialize replay buffer Replay

\For{episode in $0...M$}
 \State Reset to initial state $s_{0}$
 \For{$t = 1, T$}
  \State Sample action $a_{t} \sim \pi^{*}$; $a_{t} = b$
  \State $\tilde{a}_{t} = a_{t} +$ exploration noise
  \State Execute $\tilde{a}_{t}$ and observe reward $r_{t}$ and new state $s_{t+1}$
  \State Store transition ($s_{t}, \tilde{a}_{t}, r_{t}, s_{t+1}$) in Replay
  \State Sample a random mini-batch of transitions from replay
  \State Set $ y_{i} = r + \gamma[min(Q'(s',f'(s;\theta^{-})))]$
  \State where we sample  weights $\phi^{-}$ $S$ times and compute mean
  \State Update critic with critic loss 
  \State $L = \frac{1}{N}\sum_{i}([y_{i} - Q_{1}(s_{i},a_{i})]^2 + [y_{i} - Q_{2}(s_{i},a_{i})]^2)$
  \State where we sample  weights $\phi$ $S$ times and compute mean
  \If{episode = $N$}
    \State Update actor $\theta \gets \nabla_{\theta} Q_{1}(s,f(s;\theta))$
    \State Update the target networks
    \State $\phi^{-} \gets \tau\phi + (1-\tau)\phi^{-}$
    \State $\theta^{-} \gets \tau\theta + (1-\tau)\theta^{-}$
  \EndIf
 \EndFor
\EndFor

\end{algorithmic}
\end{algorithm}
\end{minipage}

\subsubsection{Uncertainty in states and actions}
We consider state and action space uncertainty by combining the above two algorithms. The heuristics for training the Deep Deterministic Policy Gradient (DDPG) (Supplement: Algorithm \ref{alg:nrmdp_ddpg_bayes}) and Twin Delayed DDPG (TD3) (Supplement: Algorithm \ref{alg:nrmdp_td3_bayes}) is shown in the Supplemental material.

\subsubsection{Implementation details}
In this study, the factors we optimize are as follows:
\begin{itemize}
    \item \textbf{Lockdown}: $StartDay$ and $EndDay$
    \item \textbf{Vaccination}: $StartDay$ and $EndDay$ for three age stratification.
\end{itemize}

Lockdown and vaccination days can accept any value from the start of day 0 to the end of day 100. We had two types of rewards: 
\begin{itemize}
\item{The health of the individuals ($HRew$) is given by the negative sum of mildly infected individuals and hospitalized individuals. We did not use the deceased here since our model's numbers were very low and did not include severe infections since they were all hospitalized.}
\item{The economy reward function ($ERew$) was the negative of the sum of houses with low stock.}
\end{itemize}
The total rewards ($TRew$) combine the health and the economic reward functions. A mixing factor, $\kappa$ governs the contribution of health and economic rewards. In our experiments, we use the values of $\kappa = 1$.

The policy was changed weekly in response to the states observed in the past week. The epidemic model executes the simulation using the new policy for the next week. The rewards for the RL agent are calculated at the end of the simulation time tick.

\section{Evaluation Criteria}
\label{evaluation}
As outlined previously, our study introduces two novel variants each of Deep Deterministic Policy Gradient (DDPG) and Twin Delayed Deep Deterministic Policy Gradient (TD3) reinforcement learning algorithms. Additionally, we conduct a comparative analysis, pitting the outcomes of these newly introduced algorithms against the standard versions of DDPG and TD3. To ensure a robust and replicable assessment of algorithm performance, we employ a rigorous evaluation framework structured around four distinct categories of metrics. These metrics serve as reliable indicators to assess both the performance and dependability of the algorithms. 

\subsection{Performance Metrics}
One of the majorly used performance metrics for RL algorithms has traditionally been the average reward. This is very useful in situations where other additional performance metrics cannot be derived and the rewards prove to be distinguishable between algorithms. In the epidemiological setting, we can use additional information such as the maximum peak of the infections, deaths, economic distress, etc., the variation in these produced across repeats, and the duration of the epidemic.

In this study, discovered that mean rewards do not differentiate between algorithms ``(Figure \ref{fig:overall_avg_reward})'', hence we employ the following metrics based on the epidemiological context (Equation \ref{eq:performance_metric} ) as one of the ranking metrics.

\begin{eqnarray}\label{eq:performance_metric}
metric = \sum_{I \epsilon Epi_{val}} max(I_{mean}^t) + mean(I_{max}^t - I_{min}^t) +  mean(I_{stdev}^t) \\
\end{eqnarray}

where $Epi_{val} = {Infection, Death, Hospitalization, Poverty}$.\\

\begin{figure}
  \centering
  \includegraphics[width=0.55\columnwidth]{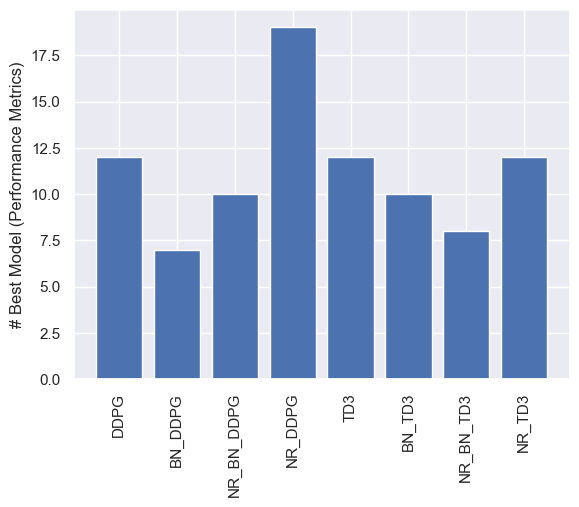}
   \caption{\textbf{Performance metrics across experiments.} T.}
  \label{fig:performance}
\end{figure}

Equation \ref{eq:performance_metric} is designed to prefer algorithms that give the lowest peak and strategies that do not give wide variations in performance. Taking the number of times each algorithm was in top 2 in terms of metrics across a range of experiments, we arrive at the best performing algorithm. This is based on the test and not training. Uncertain State and Action DDPG was the best performing algorithm ``(Figure \ref{fig:performance})''. We also observed that the in the experiments where the uncertainty was low (estimated by inducing additional noise), the performance of the vanilla algorithms was on par but with the presence of noise the novel algorithms performed better.

\subsection{Reliability Metrics}
These metrics capture the smoothness and performance drops (during training and testing) in the short and long term as a measure of reliability during training \cite{rl_reliability_metrics}.

\subsubsection{Dispersion}
We use the interquartile range (IQR) to measure dispersion and stability during training. We adopt Chan \emph{et al.}'s \cite{rl_reliability_metrics} implementation, where the rewards are detrended prior to IQR calculations. The IQR within a sliding window was chosen to keep the metric agnostic of distributions, and the detrending (i.e., $y_{t}' = y_{t} - y_{t-1})$) was done to preserve the positive increase in rewards while training. This metric was computed a few times (beginning, middle, and end) during training---the lower this value, the better the model.

\begin{figure}
  \centering
  \includegraphics[width=0.7\columnwidth]{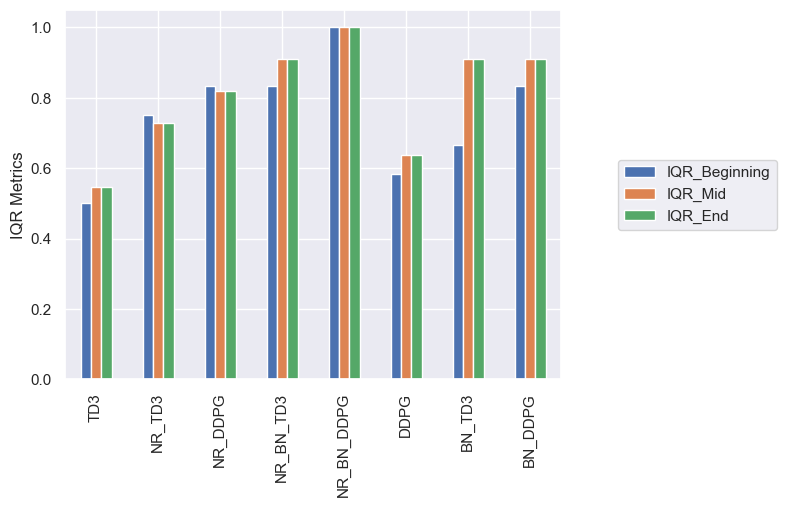}
   \caption{\textbf{Reliability Metrics measured across time.} The dispersion performance of the algorithms is given at the beginning, middle, and end stages of the training. Vanilla version of TD3 gives the lowest dispersion scores over time.}
  \label{fig:dispersion}
\end{figure}

 Vanilla TD3 was the best-performing algorithm, followed by vanilla DDPG ``(Figure \ref{fig:dispersion})''. The results show that the algorithms that account for uncertainty had higher dispersion in their rewards during training. This is probably because the uncertainty-aware models had a fluctuation in the rewards as a result of uncertainties.

\subsubsection{Risk}
We adopt the conditional value at risk (CVaR) to measure risk at training. It evaluates the expected loss value in worst-case scenarios, parameterized by a quantile $\alpha_{q}$. This gives the risk as the heaviness of the lower tail of the distribution. We compute three risk scores, Consolidated risk over time, Short-term Risk across Time (SRT), and Long-term Risk across Time (LRT).

\begin{figure}
\centering
\includegraphics[width=0.7\textwidth]{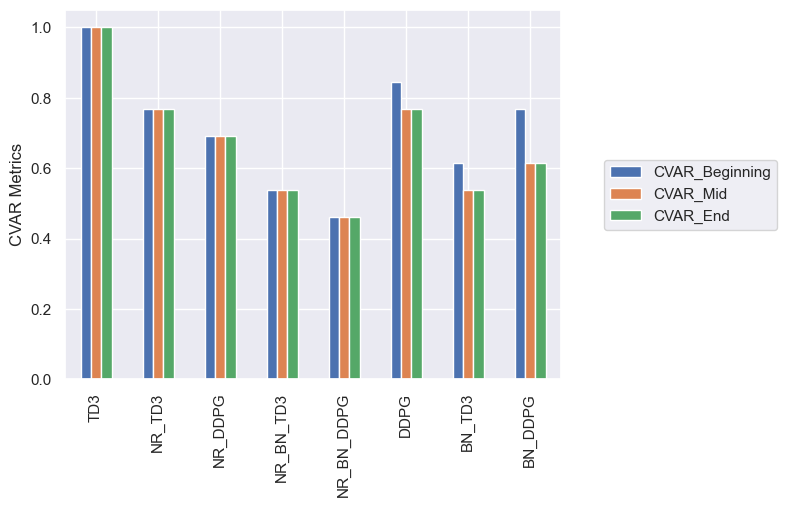} 
\caption{Consolidated Risk over Time (CVaR): Uncertain state and action TD3 is the best performing algorithm across time. refers to the short-term risk over time of the algorithms.} \label{fig:cvar}
 \end{figure}
 
\begin{figure}
\centering
\includegraphics[width=0.7\textwidth]{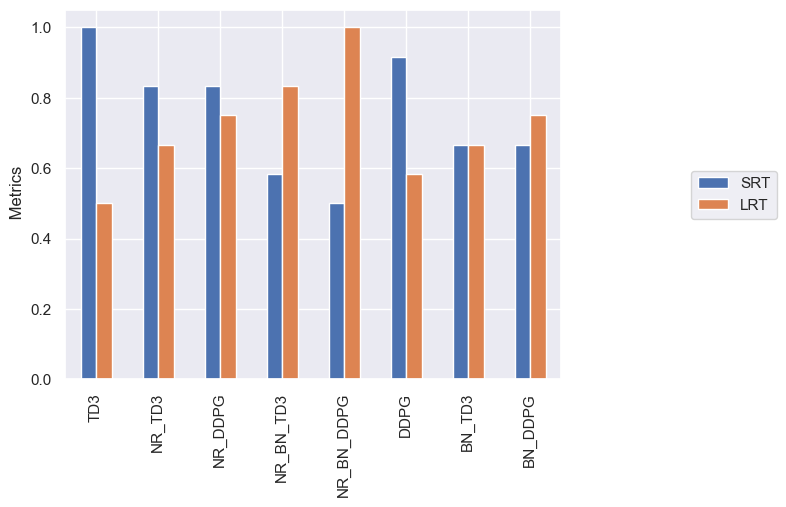} 
\caption{Short-Term and Long-term Risk over Time: On SRT, Uncertain state and action DDPG is the best performing algorithm across time and on LRT Vanilla version of TD3 gives the lowest risk scores over time.} \label{fig:srt_lrt}
\end{figure}

\begin{itemize}
   \item{\textbf{Consolidated across Time}:
   This is computed using CVaR during training at different stages (beginning, middle, and end). Lower the value, the better the performance of the algorithm. Uncertain State and Action DDPG was the best performing algorithm, followed by the Uncertain State and Action TD3 algorithm ``(Figure \ref{fig:cvar})''. The results show that the algorithms which account for uncertainty had lower overall risk.}
    \item{\textbf{Short-term Risk across Time (SRT)}: This is computed using CVaR on Differences and allows us to measure the most extreme short-term drop over time \cite{rl_reliability_metrics}. Uncertain State and Action DDPG was the best-performing algorithm, followed by Uncertain State and Action TD3 algorithms ``(Figure \ref{fig:srt_lrt})''}
    \item{\textbf{Long-term Risk across Time (LRT)}: This is computed using CVaR on Drawdown and allows us to measure the potential of the algorithm to lose a lot of performance relative to its peak over time \cite{rl_reliability_metrics}. Vanilla TD3 was the best-performing algorithm, followed by vanilla DDPG ``(Figure \ref{fig:srt_lrt})''. This work evaluated the long-term risk over a medium-long term due to the computational requirements.}
\end{itemize}

\subsection{Across Episode Trust Metrics}

A python library, ``Interestingness XRL'' \cite{Sequeira2020} calculates the trust metrics, which works for discrete state and action spaces. To extend it for continuous state and action spaces, we performed binning (Supplement: Algorithm \ref{alg:state_binning}, Supplement: Algorithm \ref{alg:action_binning}).

\subsubsection{Coverage} 
\begin{figure}
  \centering
  \includegraphics[width=0.7\columnwidth]{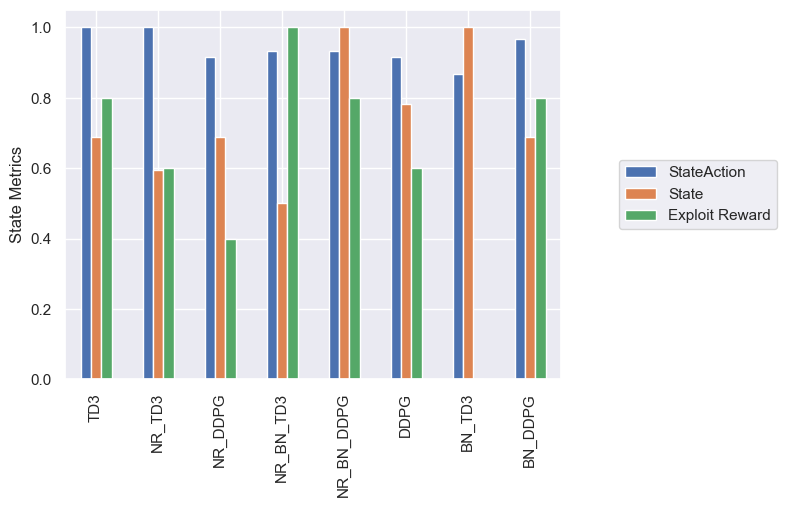}
   \caption{\textbf{Across Episode Trust Metrics.} The figure gives the state space coverage, state action coverage (during explore phase of training) and exploit rewards (during exploit phase of the training)  of the algorithms.}
   
  \label{fig:trust_metrics}
\end{figure}

Coverage gives the percent of the total states/ state-action pairs that an agent visits during training. The higher the coverage during training, the better the algorithm's performance. For a total of 750 exploration episodes, Uncertain state and action DDPG and Uncertain state and action TD3 ``(Figures \ref{fig:trust_metrics})'' covered the most state space during exploration. Uncertain state TD3 and Vanilla TD3 covered the most state-action space during exploration. The noise and the uncertainty in the model allow for better exploration of the state-space and hence better performance.

\subsubsection{Overall average reward } This metric provides us with the average reward and standard deviation to the average reward received by the agent during the training. All algorithms have similar average rewards ``(Figure \ref{fig:overall_avg_reward}'' in the Supplement). 

\subsubsection{Exploit average reward}
In addition, we also computed the rewards of the algorithms during the exploit episodes (testing). In the test plots, we could clearly observe that some algorithms had a higher average reward than the rest (forming 2 clusters). We have observed 2 clear clusters of algorithms, based on their mean rewards across all Exploit runs which are: 
    - Best-performing algorithm with mean rewards in the range of 2.6 - 3.0. 
    - Worst performing algorithm with mean rewards in the range of 1.8 - 2.4.

We computed the number of occurrences of an algorithm as the best-performing group for each of the different experiments. We rank our algorithm based on the number of times it belongs to the best-performing group (normalized) ``(Figure \ref{fig:trust_metrics})''; the higher the number of occurrences, the higher the rank. Based on our ranking we observed that Uncertain State and Action TD3 was the best-performing algorithm.

\subsection{Episode Level Trust Metrics}

\subsubsection{Reward analysis} Reward analysis discovers outliers in states and state-action pairs. It allows the policymakers to derive insights about the worst and best state-action pairs.

For some of the state-action pairs with rewards higher than the mean reward, lockdown, and vaccination start at the beginning of the simulation, which ultimately controls the spread of disease at an early stage that leads to a higher reward. In the case of state-action pairs with lower rewards, lockdown, and vaccination start at a later stage in the simulation, leading to much higher disease spread and lower reward ``(Supplement: Table \ref{tab:reward_analysis} shows State-Action outliers)''. 

\subsubsection{Transition value analysis} Transition value analysis discovers the local and absolute maxima and minima states and then computes the actions that lead to and lead out of those states. This allows the policymakers to find policies that give the best and worst-case scenarios. The result of transition value analysis across all experiments shows that there are only two absolute maxima states (which correspond to the lowest infection states).  The two states are:
\begin{itemize}
    \item \textbf{State1}: Infection\_mild- 0-33\%; Hospitalized- 0-33\%; House with a minimum stock: 0-33\%
    \item \textbf{State2}: Infection\_mild- 0-33\%; Hospitalized- 0-33\%; House with a minimum stock: 66-100\%
\end{itemize}

\begin{figure}
  \centering
  \includegraphics[width=0.55\columnwidth]{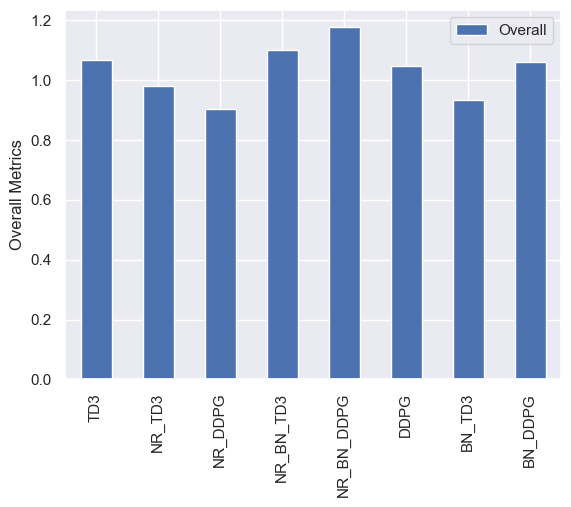}
   \caption{\textbf{Consolidated Metric.} Uncertain State and Action DDPG was the best performaing algorithm followed by Uncertain State and Action TD3.}
   
  \label{fig:overall}
\end{figure}
 
 Each of these scores evaluate different facets of the algorithms and there is no single algorithm which does well on all metrics. Hence, we compute a 
consolidated overall score as a combination of all the numerical scores computed above. Uncertain State and Action TD3 was the best-performing algorithm, followed by Uncertain State and Action DDPG and Uncertain State TD3 algorithms ``(Figure \ref{fig:overall})''. The results show that the algorithms that account for the uncertainty during their training are more reliable.

 When we compared the different algorithms, using metrics and the control of epidemics, uncertainty-aware state, and action methods were among the best-performing models. Since the training of DDPG and TD3 is essentially noise driven, given the same noise function and the environment, the algorithms that had a relatively more noisy Q value, the actions and states could explore better than the vanilla version of the algorithms. As a result, the uncertainty-aware models performed better than the vanilla versions of the algorithms.

\section{Experiments} \label{experiments}
The epidemic model discussed in ``(Sec \ref{sec:epi-model})'' is executed with 1000 agents for 100 days, i.e., 600 simulation ticks. The policy is updated every seven days, i.e., every 42 simulation ticks. Our experiments have two types of vaccines with 80\% and 60\% effectiveness and two types of masks with 80\% and 40\% effectiveness. The different experiments we tested are as follows ``(Table \ref{tab:vac-mask-availability})'': All the experiments are repeated for DDPG, and TD3 algorithms with \textit{Uncertainty in Actions} (NR\_DDPG, NR\_TD3), \textit{Uncertainty in States}(BN\_DDPG, BN\_TD3) and \textit{Uncertainty in States and Actions}(NR\_BN\_DDPG, NR\_BN\_TD3) discussed in ``(Sec \ref{sec:public_policy})''.

    \begin{table}[htbp]
        \caption{Vaccines and Masks Availability For Experiments}
        \begin{tabular}{|c|c|c|c|c|}
        \toprule
        \textbf{Name} & \textbf{Vaccine 1 }& \textbf{Vaccine 2}  & \textbf{Mask 1} & \textbf{Mask 2} \\
        \midrule
         Baseline & 6 & 6 & 500 &  1000  \\
         High Mask & 6 & 6 & 800 & 1000  \\
         Low Mask & 6 & 6 & 100 & 1000  \\
         High Vaccine & 12 & 6 & 500 & 1000 \\  
         Low Vaccine & 6 & 12 & 500 & 1000  \\
         \bottomrule
     \multicolumn{5}{l}{\textbf{*Vaccines doses available per day}}
        \end{tabular}
        \label{tab:vac-mask-availability}
    \end{table}

We evaluate each experiment for different levels of initial infections, (1\%, 5\%, and 10\%). We also simulate uncertainty in the measurement of the states and actions by sampling from a noisy distribution with noise level, (0, 3\%, and 5\%). For each setting, we repeat the experiments three times. In all of these experiments, we start with a default lockdown setting to emulate a scenario where the default disaster response is triggered prior to the informed response made by the policymakers later in the epidemic. The algorithm's response is set as policy after 1 week into the epidemic simulation.

\subsection{Baseline Experiments}
We tested a baseline scenario where everyone can get vaccinated over 100 days should they choose to get vaccinated. Similarly, everyone can get a low-efficiency mask, and up to 50\% people can get a high-efficiency one should they decide to get it. 

\begin{figure*}
  \centering
  \caption{Baseline Experiment: Uncertain state TD3 was the best-performing model (performance metrics). Early interventions help control the epidemic faster.}

    \begin{subfigure}[ht]{\textwidth}
    \centering
    \caption{Epidemiological Curves by Initial Infection \% ($Inf$) and Uncertainty in measurements ($Noise$)}
    \hspace*{-1.75cm}
    \includegraphics[width=1.4\textwidth]{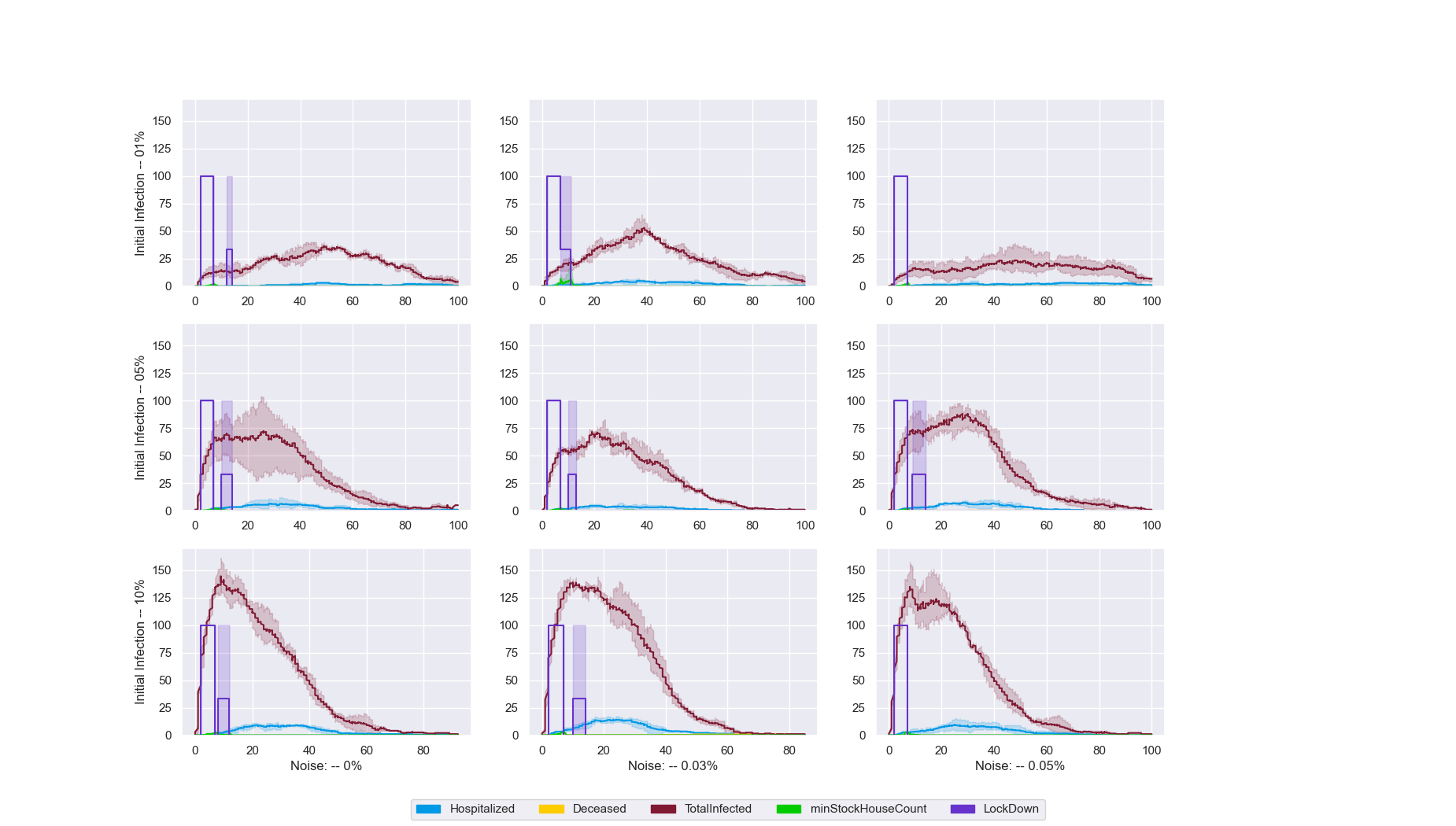} \label{fig:baseline_epi}
  \end{subfigure}
  
  \begin{subtable}[t]{\textwidth}
    \centering
    \caption{Actions suggested by the model with respect to Initial Infection \% ($Inf$) and Uncertainty in measurements ($Noise$)}
    \label{tab:baseline_actions}
    \begin{tabular}{|r|r|l|l|l|l|l|l|} 
    \toprule
    Inf & Noise & $V^{0-17}_{Start}$ & $V^{0-17}_{End}$ & $V^{18-59}_{Start}$ & $V^{18-59}_{End}$ & $V^{60-99}_{Start}$ & $V^{60-99}_{End}$ \\
    \midrule
    0.01 & 0.00 & 3.7 $\pm$ 0.1 & 3.7 $\pm$ 0.2 & 0.2 $\pm$ 0.7 & 2.5 $\pm$ 0.6 & 3.1 $\pm$ 1.1 & 9.3 $\pm$ 0.2 \\
    0.01 & 0.03 & 3.6 $\pm$ 0.2 & 3.6 $\pm$ 0.1 & 0.3 $\pm$ 1.0 & 2.6 $\pm$ 0.8 & 3.0 $\pm$ 0.7 & 9.1 $\pm$ 0.9 \\
    0.01 & 0.05 & 3.6 $\pm$ 0.0 & 3.6 $\pm$ 0.0 & 0.2 $\pm$ 0.6 & 2.4 $\pm$ 0.1 & 3.1 $\pm$ 1.0 & 9.3 $\pm$ 0.3 \\
    0.05 & 0.00 & 3.6 $\pm$ 0.0 & 3.7 $\pm$ 0.1 & 0.2 $\pm$ 0.7 & 2.5 $\pm$ 0.6 & 3.1 $\pm$ 1.0 & 9.2 $\pm$ 0.7 \\
    0.05 & 0.03 & 3.5 $\pm$ 0.3 & 3.5 $\pm$ 0.3 & 0.1 $\pm$ 0.5 & 2.4 $\pm$ 0.1 & 3.0 $\pm$ 0.5 & 9.2 $\pm$ 0.6 \\
    0.05 & 0.05 & 3.6 $\pm$ 0.0 & 3.8 $\pm$ 0.6 & 0.3 $\pm$ 0.9 & 2.5 $\pm$ 0.7 & 3.1 $\pm$ 1.0 & 9.1 $\pm$ 0.6 \\
    0.10 & 0.00 & 3.6 $\pm$ 0.0 & 3.7 $\pm$ 0.2 & 0.2 $\pm$ 0.8 & 2.4 $\pm$ 0.3 & 3.1 $\pm$ 0.9 & 9.3 $\pm$ 0.3 \\
    0.10 & 0.03 & 3.7 $\pm$ 0.1 & 3.8 $\pm$ 0.6 & 0.2 $\pm$ 0.9 & 2.6 $\pm$ 0.8 & 3.1 $\pm$ 1.0 & 9.3 $\pm$ 0.2 \\
    0.10 & 0.05 & 3.6 $\pm$ 0.0 & 3.7 $\pm$ 0.1 & 0.3 $\pm$ 1.1 & 2.6 $\pm$ 1.1 & 3.1 $\pm$ 0.9 & 9.2 $\pm$ 0.4 \\
    \bottomrule
    \end{tabular}
  \end{subtable}
\end{figure*}

\begin{figure}[!htb]
  \centering
   \hspace*{-1.75cm}
  \includegraphics[width=1.4\textwidth]{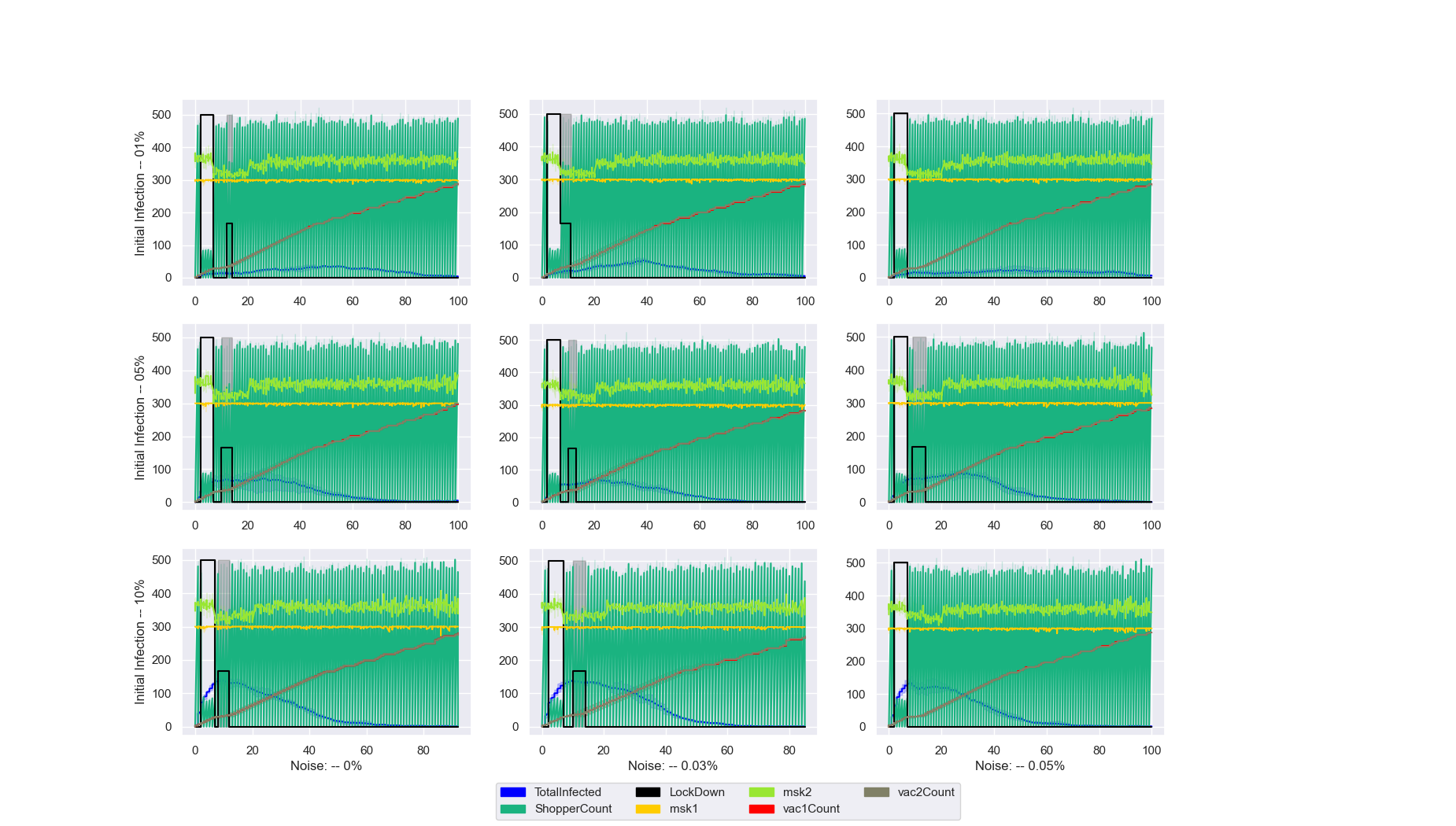}
   \caption{\textbf{Baseline Experiment} Detailed masking, vaccination, shopping and lockdown behavior}
  \label{fig:baseline_appendix}
\end{figure}

The purpose of the baseline experiment was to check if the algorithms presented so far could effectively control the epidemic. All algorithms could control the outbreak to an extent ``(Figure \ref{fig:baseline_epi}, Table \ref{tab:baseline_actions})''. However, uncertain state TD3 was the best-performing model that could effectively control the infections and stabilize the economy. The peak infection was capped between 25-50 individuals at 1\% initial infections, 50-90 for 5\%, and approximately 150  for 10\% initial infections.  In all the experiments, the economy was also stabilized in the initial days of the epidemic. About 70\% of the population wore masks, where the high-efficiency mask levels remained stable throughout the experiment and the low-efficiency mask levels fluctuated based on the requirement. For example, people wore more masks during lockdowns ``(Figure \ref{fig:baseline_appendix})'' and shopped less as opposed to no lockdown times. The vaccination was carried out at a constant maximum rate with both vaccines. The algorithm preferred to vaccinate the most mobile population (18-59 years), followed by the older population (61-90), and finally the younger population. The older populations (60-99) had the longest time window for vaccination (3rd day onwards), and the youngest (0-17) had the lowest vaccination time window (less than 1 day in a week). With respect to this baseline, we tested other scenarios to determine the role of masks, especially high-efficiency masks, lockdowns, and vaccinations, in controlling public health and stabilizing the economy.

\subsection{High-Mask Experiments}
In this experiment, we wanted to check if people chose to use high-efficiency masks if they were made available in relatively higher numbers (up to 80\% of the population) and see if mask availability could control the epidemic better than baseline. Therefore, compared to the baseline, we varied the high-efficiency mask availability to up to 80\% of the population. 

\begin{figure*}
  \centering
  \caption{High Mask Experiment: Uncertain action DDPG was the best-performing model (performance metrics). Early interventions help control the epidemic faster.}

    \begin{subfigure}[t]{\textwidth}
    \centering
    \caption{Epidemiological Curves by Initial Infection \% ($Inf$) and Uncertainty in measurements ($Noise$)}
    \hspace*{-1.75cm}
    \includegraphics[width=1.4\textwidth]{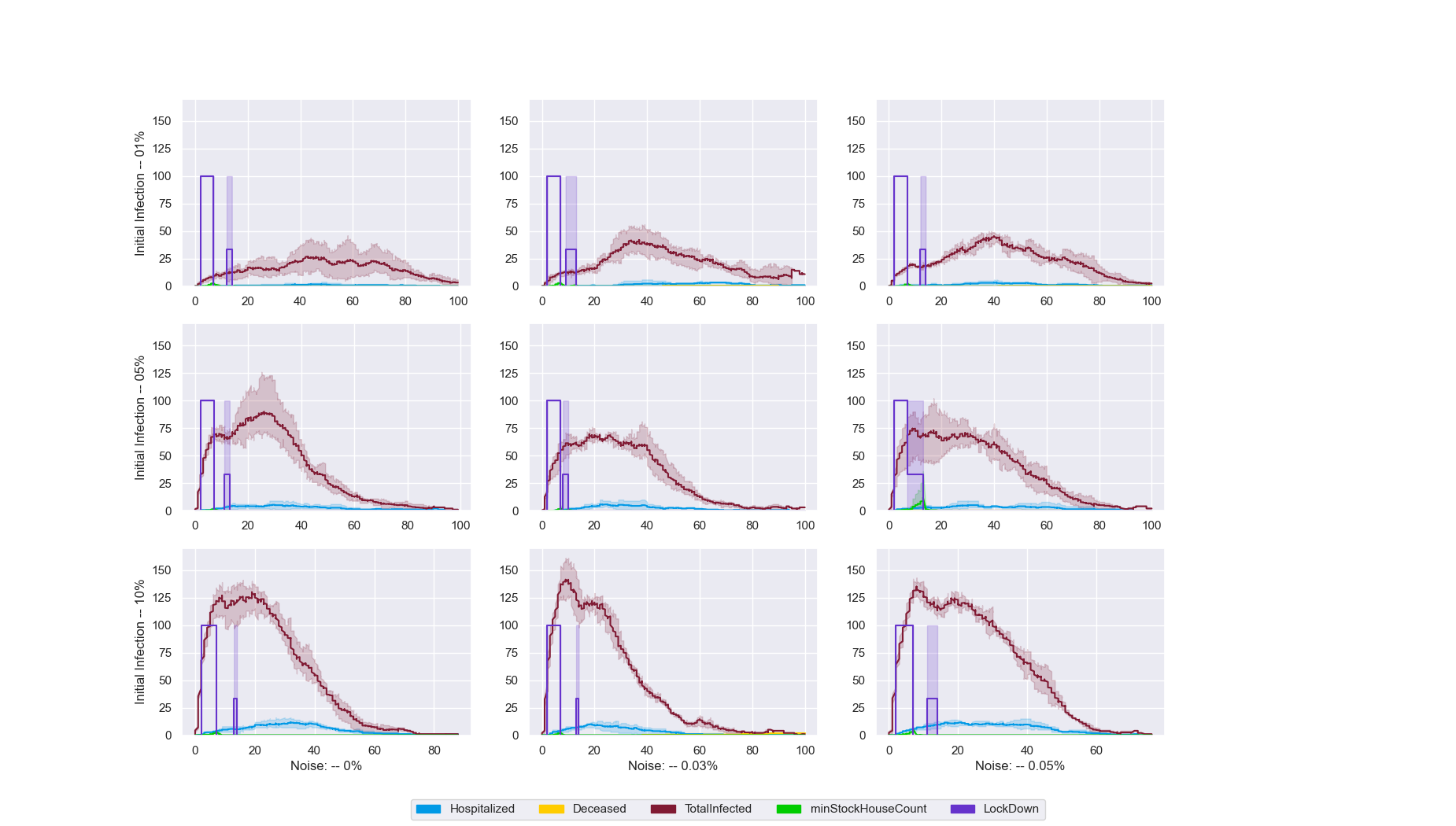} \label{fig:highmask_epi}
  \end{subfigure}
  
  \begin{subtable}[t]{\textwidth}
    \centering
    \caption{Actions suggested by the model with respect to Initial Infection \% ($Inf$) and Uncertainty in measurements ($Noise$)}
    \label{tab:highmask_actions}
    \begin{tabular}{|r|r|l|l|l|l|l|l|}
    \toprule
    Inf & Noise & $V^{0-17}_{Start}$ & $V^{0-17}_{End}$ & $V^{18-59}_{Start}$ & $V^{18-59}_{End}$ & $V^{60-99}_{Start}$ & $V^{60-99}_{End}$ \\
    \midrule
    0.01 & 0.00 & 4.0 $\pm$ 0.2 & 4.0 $\pm$ 0.2 & 0.3 $\pm$ 1.0 & 2.6 $\pm$ 0.6 & 3.3 $\pm$ 0.9 & 9.6 $\pm$ 0.4 \\
    0.01 & 0.03 & 3.8 $\pm$ 0.2 & 4.0 $\pm$ 0.2 & 0.2 $\pm$ 0.9 & 2.6 $\pm$ 0.8 & 3.2 $\pm$ 0.6 & 9.5 $\pm$ 0.7 \\
    0.01 & 0.05 & 3.9 $\pm$ 0.1 & 3.9 $\pm$ 0.1 & 0.2 $\pm$ 0.7 & 2.5 $\pm$ 0.2 & 3.2 $\pm$ 0.5 & 9.5 $\pm$ 0.8 \\
    0.05 & 0.00 & 3.9 $\pm$ 0.1 & 3.9 $\pm$ 0.1 & 0.2 $\pm$ 0.6 & 2.5 $\pm$ 0.4 & 3.3 $\pm$ 1.0 & 9.5 $\pm$ 0.7 \\
    0.05 & 0.03 & 3.9 $\pm$ 0.2 & 4.0 $\pm$ 0.4 & 0.3 $\pm$ 1.1 & 2.7 $\pm$ 1.0 & 3.2 $\pm$ 0.6 & 9.6 $\pm$ 0.6 \\
    0.05 & 0.05 & 3.9 $\pm$ 0.2 & 3.9 $\pm$ 0.0 & 0.3 $\pm$ 0.9 & 2.6 $\pm$ 0.7 & 3.2 $\pm$ 0.7 & 9.6 $\pm$ 0.4 \\
    0.10 & 0.00 & 3.9 $\pm$ 0.1 & 4.1 $\pm$ 0.6 & 0.3 $\pm$ 0.9 & 2.7 $\pm$ 0.9 & 3.3 $\pm$ 1.0 & 9.6 $\pm$ 0.5 \\
    0.10 & 0.03 & 3.8 $\pm$ 0.3 & 3.9 $\pm$ 0.1 & 0.2 $\pm$ 0.7 & 2.5 $\pm$ 0.2 & 3.3 $\pm$ 1.0 & 9.7 $\pm$ 0.2 \\
    0.10 & 0.05 & 3.8 $\pm$ 0.2 & 3.9 $\pm$ 0.1 & 0.3 $\pm$ 1.0 & 2.6 $\pm$ 0.5 & 3.1 $\pm$ 0.6 & 9.4 $\pm$ 1.0 \\
    \bottomrule
    \end{tabular}
  \end{subtable}
\end{figure*}

Uncertain action DDPG was the best-performing model that could effectively control both the infections and the economy ``(Figure \ref{fig:highmask_epi}, Table \ref{tab:highmask_actions})''. The infection peak was very similar to the baseline scenario when the initial infection and the uncertainty in the state measurement were low. This was because the high-efficiency masks were very well utilized (nearly 30\% of the population). As the initial infection increased and the information flow became chaotic due to higher noise/ uncertainty, the infections were lower and the peak flattened when better masks were made available. Similar to the baseline, 70\% of the population wore masks and people shopped less during lockdowns as opposed to no lockdown times ``(Figure \ref{fig:highmask_appendix})''. The difference between the baseline is that start dates for the older and younger populations were delayed slightly (0.2 days) on average as opposed to the baseline.

\subsection{Low-Mask Experiments}
In this experiment, we wanted to check the response when high-efficiency masks became scarce (up to 20\% of the population) and see if the epidemic got worse than the baseline. 

\begin{figure*}
  \centering
  \caption{Low Mask Experiment: Uncertain action TD3 was the best-performing model (performance metrics). Early interventions help control the epidemic faster.}

    \begin{subfigure}[t]{\textwidth}
    \centering
    \caption{Epidemiological Curves by Initial Infection \% ($Inf$) and Uncertainty in measurements ($Noise$)}
    \hspace*{-1.75cm}
    \includegraphics[width=1.4\textwidth]{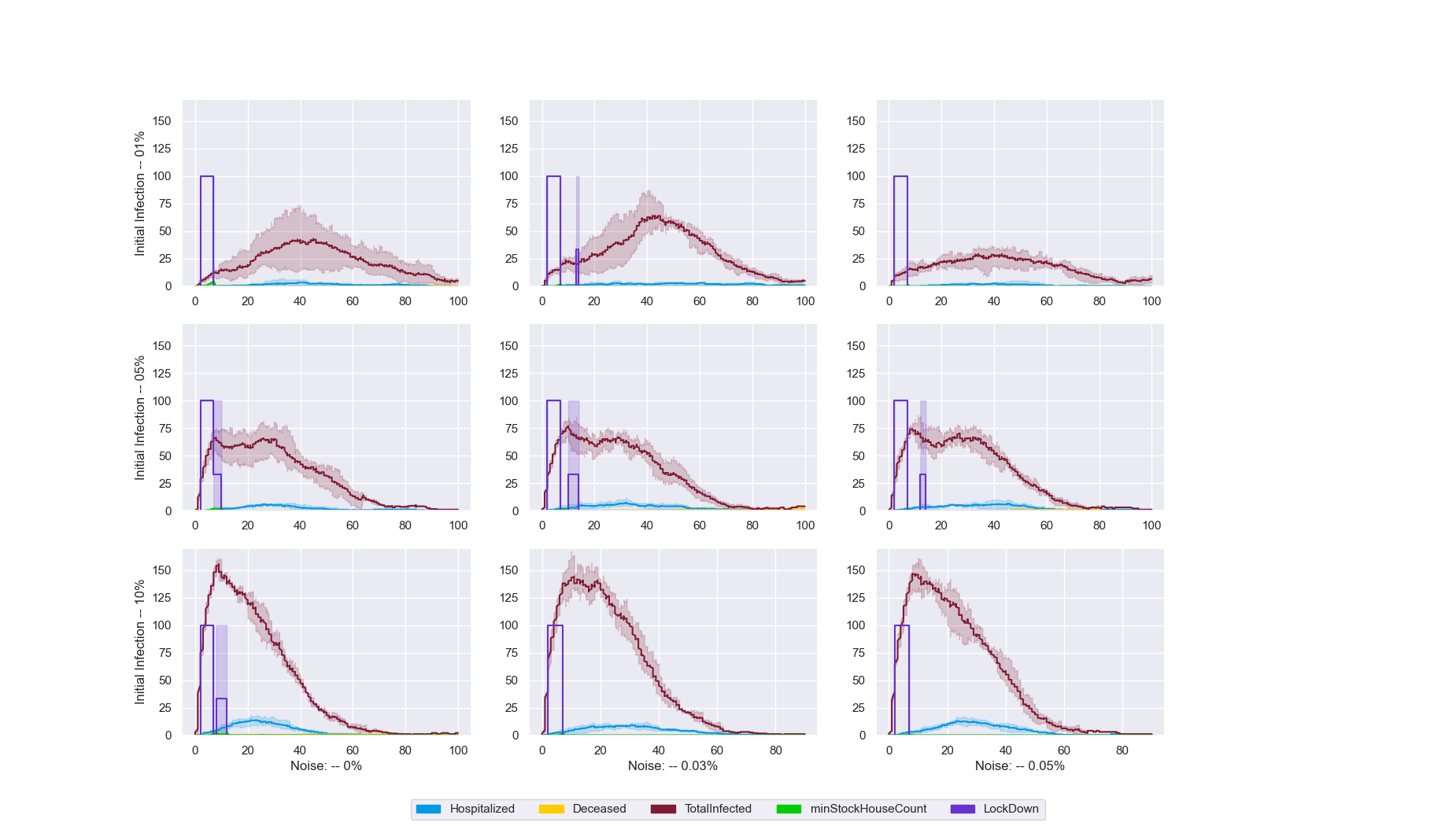} \label{fig:lowmask_epi}
  \end{subfigure}
  
  \begin{subtable}[t]{\textwidth}
    \centering
    \caption{Actions suggested by the model with respect to Initial Infection \% ($Inf$) and Uncertainty in measurements ($Noise$)}
    \label{tab:lowmask_actions}
    \begin{tabular}{|r|r|l|l|l|l|l|l|}
    \toprule
    Inf & Noise & $V^{0-17}_{Start}$ & $V^{0-17}_{End}$ & $V^{18-59}_{Start}$ & $V^{18-59}_{End}$ & $V^{60-99}_{Start}$ & $V^{60-99}_{End}$ \\
    \midrule
    0.01 & 0.00 & 3.7 $\pm$ 0.3 & 3.9 $\pm$ 0.8 & 0.3 $\pm$ 1.1 & 2.5 $\pm$ 0.6 & 3.1 $\pm$ 0.6 & 9.2 $\pm$ 0.7 \\
    0.01 & 0.03 & 3.7 $\pm$ 0.2 & 3.8 $\pm$ 0.6 & 0.3 $\pm$ 1.1 & 2.6 $\pm$ 1.0 & 3.2 $\pm$ 1.0 & 9.3 $\pm$ 0.4 \\
    0.01 & 0.05 & 3.7 $\pm$ 0.1 & 3.7 $\pm$ 0.2 & 0.2 $\pm$ 0.9 & 2.5 $\pm$ 0.6 & 3.1 $\pm$ 1.0 & 9.3 $\pm$ 0.3 \\
    0.05 & 0.00 & 3.6 $\pm$ 0.1 & 3.7 $\pm$ 0.3 & 0.3 $\pm$ 1.0 & 2.5 $\pm$ 0.6 & 3.2 $\pm$ 1.0 & 9.2 $\pm$ 0.6 \\
    0.05 & 0.03 & 3.6 $\pm$ 0.2 & 3.7 $\pm$ 0.2 & 0.2 $\pm$ 0.8 & 2.5 $\pm$ 0.7 & 3.2 $\pm$ 1.0 & 9.4 $\pm$ 0.1 \\
    0.05 & 0.05 & 3.6 $\pm$ 0.2 & 3.6 $\pm$ 0.2 & 0.2 $\pm$ 0.7 & 2.5 $\pm$ 0.6 & 3.0 $\pm$ 0.5 & 9.1 $\pm$ 1.1 \\
    0.10 & 0.00 & 3.6 $\pm$ 0.2 & 3.7 $\pm$ 0.3 & 0.2 $\pm$ 0.8 & 2.5 $\pm$ 0.7 & 3.0 $\pm$ 0.5 & 9.1 $\pm$ 0.7 \\
    0.10 & 0.03 & 3.6 $\pm$ 0.1 & 3.7 $\pm$ 0.3 & 0.3 $\pm$ 1.1 & 2.6 $\pm$ 0.8 & 3.0 $\pm$ 0.6 & 9.1 $\pm$ 0.9 \\
    0.10 & 0.05 & 3.7 $\pm$ 0.1 & 3.8 $\pm$ 0.6 & 0.2 $\pm$ 0.8 & 2.4 $\pm$ 0.2 & 3.1 $\pm$ 0.7 & 9.1 $\pm$ 0.9 \\
    \bottomrule
    \end{tabular}
  \end{subtable}
\end{figure*}

\begin{figure}
  \centering
  \hspace*{-1.75cm}
  \includegraphics[width=1.4\textwidth]{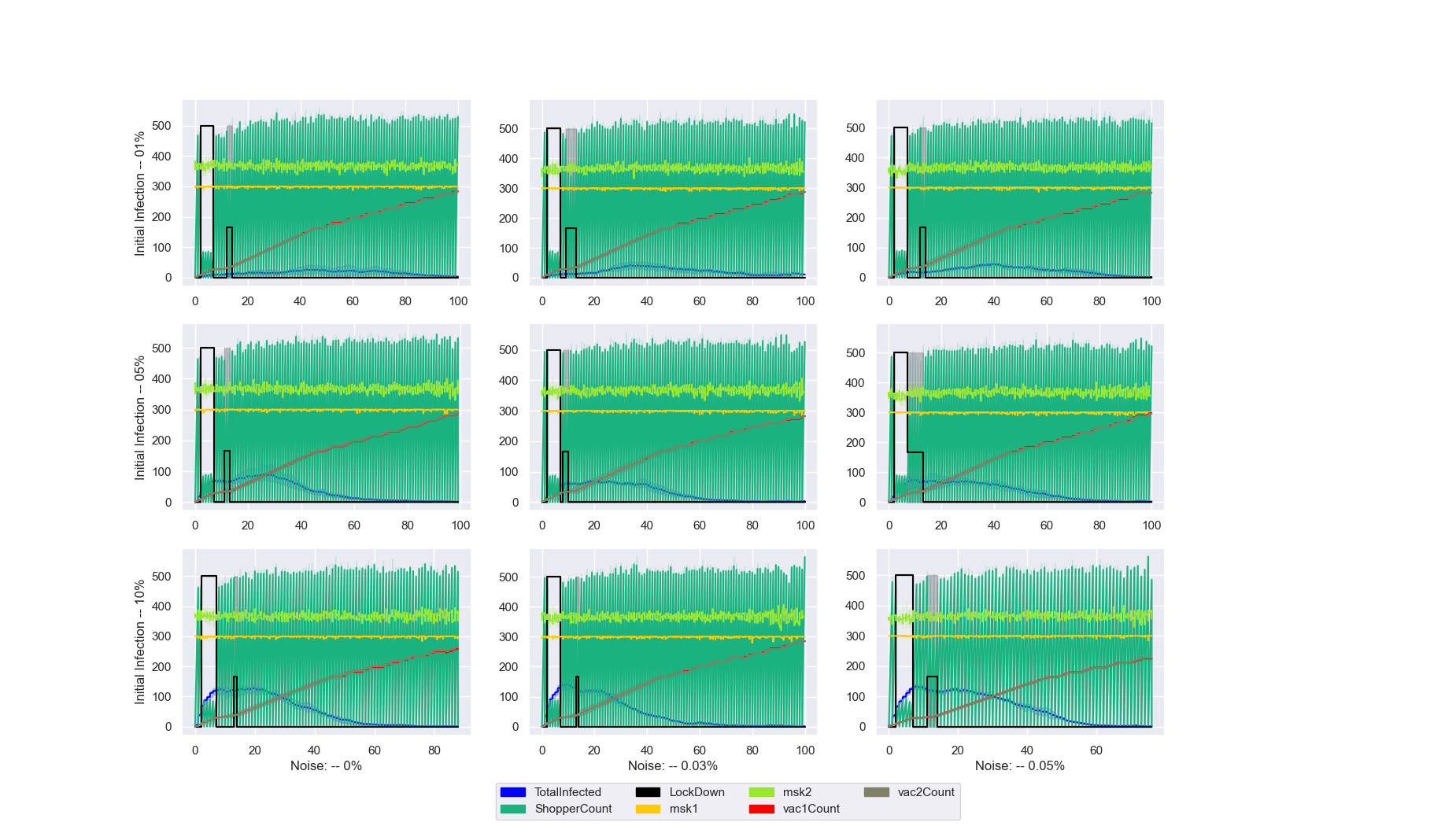}
   \caption{\textbf{High Mask Experiment} Detailed masking, vaccination, shopping and lockdown behavior}
  \label{fig:highmask_appendix}
\end{figure}

Uncertain action TD3 was the best-performing model that could effectively control both the infections and the economy ``(Figure \ref{fig:lowmask_epi}, Table \ref{tab:lowmask_actions})''. The variance in the infection peak is higher than both the baseline and high mask. In addition, with the higher uncertainty and initial infections, when there are low-quality masks available the peak is higher as compared to baseline and high-mask experiments. There is also a left shift of the peak in the low mask experiment for the total infections and hospitalizations. These results demonstrate that the availability of high-efficiency masks can help control the epidemic effectively and help flatten the curve to control hospitalizations at any point in time. 70\% of the population wore masks, and vaccinations occurred similar to the baseline ``(Figure \ref{fig:highmask_appendix})''. The only caveat is that the rate of vaccinations slowed down slightly after the 40-day mark maybe because it coincides with the fall of infections and a majority of the population had already recovered or had been vaccinated.

\subsection{High-Vaccine Experiments}
In this experiment, we wanted to check the effect of high-efficiency vaccines if they were made available in relatively higher numbers (up to 100\% of the population). 

\begin{figure*}
  \centering
  \caption{High Vaccine Experiment: Uncertain state-action DDPG was the best-performing model (performance metrics). Early interventions help control the epidemic faster.}

    \begin{subfigure}[t]{\textwidth}
    \centering
    \caption{Epidemiological Curves by Initial Infection \% ($Inf$) and Uncertainty in measurements ($Noise$)}
    \hspace*{-1.75cm}
    \includegraphics[width=1.4\textwidth]{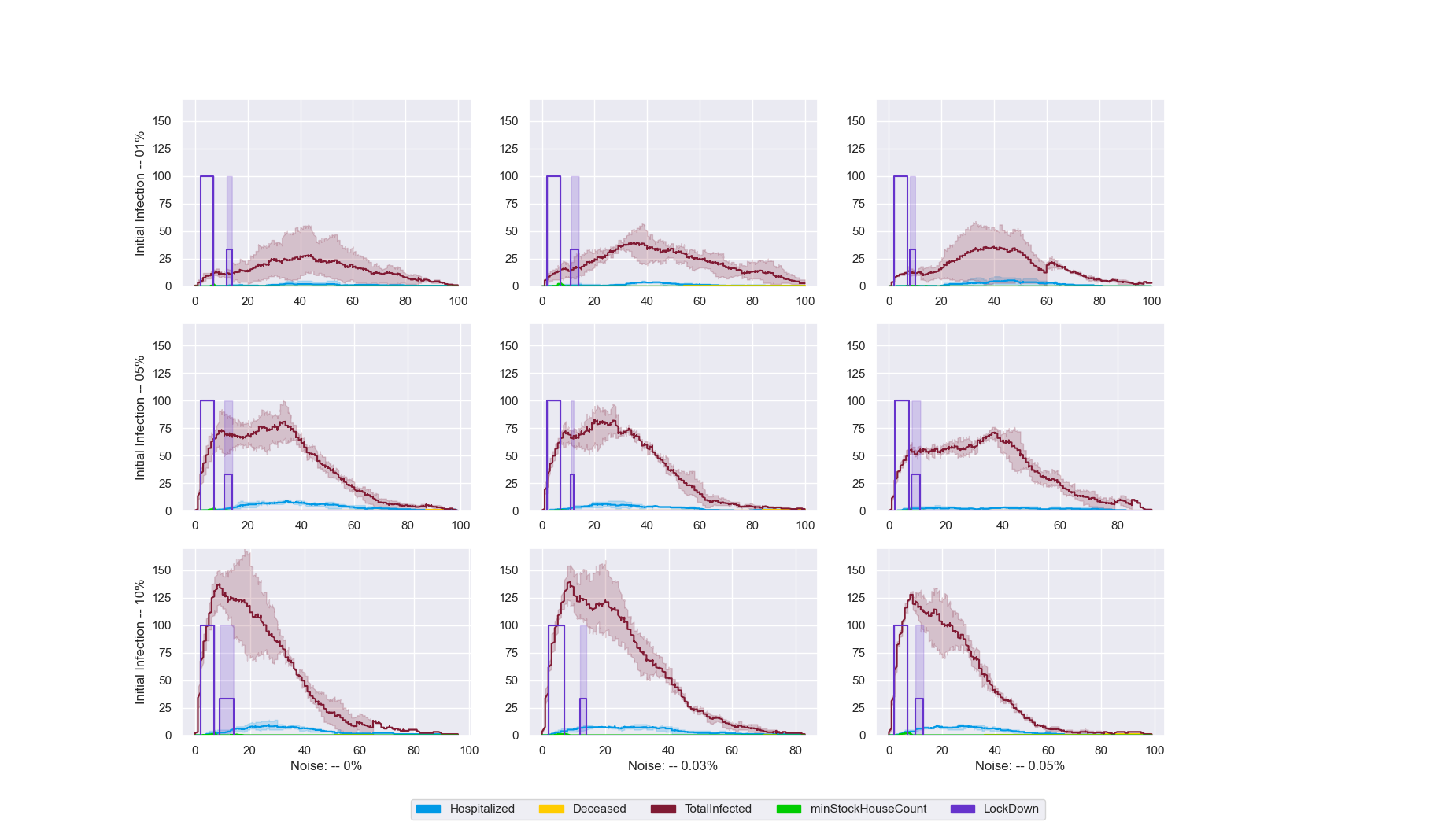} \label{fig:highvaccine_epi}
  \end{subfigure}
  
  \begin{subtable}[t]{\textwidth}
    \centering
    \caption{Actions suggested by the model with respect to Initial Infection \% ($Inf$) and Uncertainty in measurements ($Noise$)}
    \label{tab:highvaccine_actions}
    \begin{tabular}{|r|r|l|l|l|l|l|l|}
    \toprule
    Inf & Noise & $V^{0-17}_{Start}$ & $V^{0-17}_{End}$ & $V^{18-59}_{Start}$ & $V^{18-59}_{End}$ & $V^{60-99}_{Start}$ & $V^{60-99}_{End}$ \\
    \midrule
    0.01 & 0.00 & 3.7 $\pm$ 0.2 & 3.8 $\pm$ 0.3 & 0.2 $\pm$ 0.7 & 2.4 $\pm$ 0.4 & 3.4 $\pm$ 0.4 & 9.3 $\pm$ 1.0 \\
    0.01 & 0.03 & 3.7 $\pm$ 0.3 & 3.7 $\pm$ 0.0 & 0.2 $\pm$ 0.6 & 2.4 $\pm$ 0.2 & 3.4 $\pm$ 0.6 & 9.4 $\pm$ 0.7 \\
    0.01 & 0.05 & 3.7 $\pm$ 0.3 & 3.8 $\pm$ 0.1 & 0.2 $\pm$ 0.6 & 2.5 $\pm$ 0.6 & 3.4 $\pm$ 0.4 & 9.3 $\pm$ 0.9 \\
    0.05 & 0.00 & 3.7 $\pm$ 0.2 & 3.9 $\pm$ 0.4 & 0.3 $\pm$ 1.1 & 2.5 $\pm$ 0.7 & 3.5 $\pm$ 0.6 & 9.3 $\pm$ 1.0 \\
    0.05 & 0.03 & 3.7 $\pm$ 0.1 & 3.8 $\pm$ 0.3 & 0.2 $\pm$ 0.8 & 2.5 $\pm$ 0.5 & 3.4 $\pm$ 0.6 & 9.3 $\pm$ 1.0 \\
    0.05 & 0.05 & 3.7 $\pm$ 0.2 & 3.8 $\pm$ 0.1 & 0.3 $\pm$ 1.0 & 2.4 $\pm$ 0.4 & 3.4 $\pm$ 0.5 & 9.3 $\pm$ 0.9 \\
    0.10 & 0.00 & 3.7 $\pm$ 0.1 & 3.8 $\pm$ 0.2 & 0.2 $\pm$ 0.6 & 2.4 $\pm$ 0.4 & 3.4 $\pm$ 0.5 & 9.3 $\pm$ 1.0 \\
    0.10 & 0.03 & 3.7 $\pm$ 0.1 & 3.9 $\pm$ 0.4 & 0.2 $\pm$ 0.8 & 2.5 $\pm$ 0.7 & 3.4 $\pm$ 0.5 & 9.4 $\pm$ 0.8 \\
    0.10 & 0.05 & 3.7 $\pm$ 0.1 & 3.8 $\pm$ 0.0 & 0.2 $\pm$ 0.9 & 2.5 $\pm$ 0.5 & 3.4 $\pm$ 0.5 & 9.4 $\pm$ 0.6 \\
    \bottomrule
    \end{tabular}
  \end{subtable}
\end{figure*}

Uncertain state-action DDPG was the best-performing model that could effectively control both the infections and the economy ``(Figure \ref{fig:highvaccine_epi}, Table \ref{tab:highvaccine_actions})''. The peak infection in the best-performing model was similar to the baseline scenario for high infection scenarios. For the lower infection, the peak and hospitalizations were slightly lower than the baseline. In the high initial infection scenarios, the vaccination had no discernable effect over the baseline. The vaccination rates were at the maximum allowed till the 50 days following which it slowed down slightly as more people got vaccinated or recovered ``(Figure \ref{fig:highvaccine_appendix})''.

\subsection{Low-Vaccine Experiments}
\begin{figure*}
  \centering
  \caption{Low Vaccine Experiment: Uncertain action DDPG was the best-performing model (performance metrics). Early interventions help control the epidemic faster.}

    \begin{subfigure}[t]{\textwidth}
    \centering
    \caption{Epidemiological Curves by Initial Infection \% ($Inf$) and Uncertainty in measurements ($Noise$)}
    \hspace*{-1.75cm}
    \includegraphics[width=1.4\textwidth]{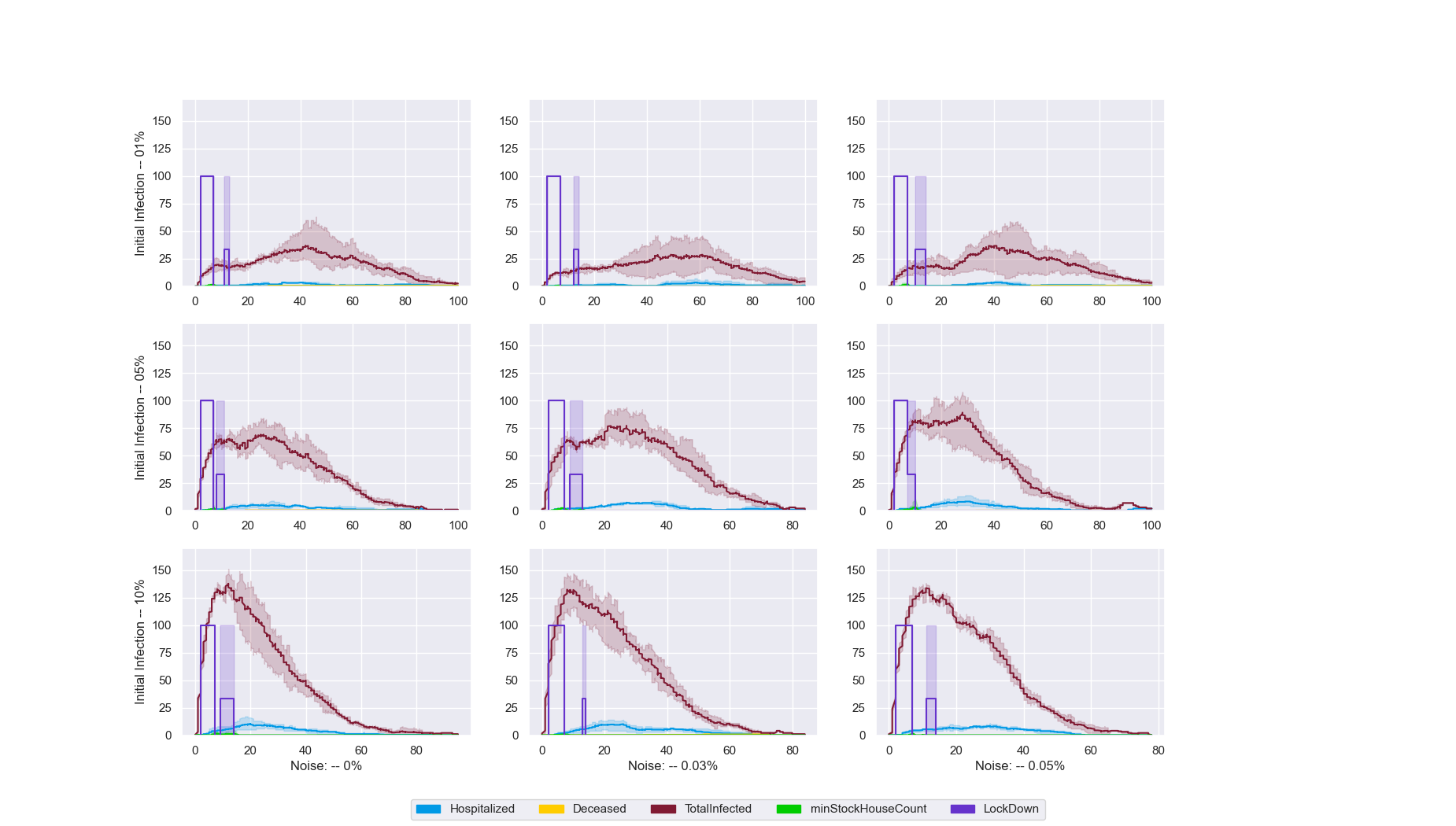} \label{fig:lowvaccine_epi}
  \end{subfigure}
  
  \begin{subtable}[t]{\textwidth}
    \centering
    \caption{Actions suggested by the model with respect to Initial Infection \% ($Inf$) and Uncertainty in measurements ($Noise$)}
    \label{tab:lowvaccine_actions}
    \begin{tabular}{|r|r|l|l|l|l|l|l|}
    \toprule
    Inf & Noise & $V^{0-17}_{Start}$ & $V^{0-17}_{End}$ & $V^{18-59}_{Start}$ & $V^{18-59}_{End}$ & $V^{60-99}_{Start}$ & $V^{60-99}_{End}$ \\
    \midrule
    0.01 & 0.00 & 4.0 $\pm$ 0.2 & 4.0 $\pm$ 0.2 & 0.3 $\pm$ 1.0 & 2.6 $\pm$ 0.6 & 3.3 $\pm$ 0.9 & 9.6 $\pm$ 0.4 \\
    0.01 & 0.03 & 3.8 $\pm$ 0.2 & 4.0 $\pm$ 0.2 & 0.2 $\pm$ 0.9 & 2.6 $\pm$ 0.8 & 3.2 $\pm$ 0.6 & 9.5 $\pm$ 0.7 \\
    0.01 & 0.05 & 3.9 $\pm$ 0.1 & 3.9 $\pm$ 0.1 & 0.2 $\pm$ 0.7 & 2.5 $\pm$ 0.2 & 3.2 $\pm$ 0.5 & 9.5 $\pm$ 0.8 \\
    0.05 & 0.00 & 3.9 $\pm$ 0.1 & 3.9 $\pm$ 0.1 & 0.2 $\pm$ 0.6 & 2.5 $\pm$ 0.4 & 3.3 $\pm$ 1.0 & 9.5 $\pm$ 0.7 \\
    0.05 & 0.03 & 3.9 $\pm$ 0.2 & 4.0 $\pm$ 0.4 & 0.3 $\pm$ 1.1 & 2.7 $\pm$ 1.0 & 3.2 $\pm$ 0.6 & 9.6 $\pm$ 0.6 \\
    0.05 & 0.05 & 3.9 $\pm$ 0.2 & 3.9 $\pm$ 0.0 & 0.3 $\pm$ 0.9 & 2.6 $\pm$ 0.7 & 3.2 $\pm$ 0.7 & 9.6 $\pm$ 0.4 \\
    0.10 & 0.00 & 3.9 $\pm$ 0.1 & 4.1 $\pm$ 0.6 & 0.3 $\pm$ 0.9 & 2.7 $\pm$ 0.9 & 3.3 $\pm$ 1.0 & 9.6 $\pm$ 0.5 \\
    0.10 & 0.03 & 3.8 $\pm$ 0.3 & 3.9 $\pm$ 0.1 & 0.2 $\pm$ 0.7 & 2.5 $\pm$ 0.2 & 3.3 $\pm$ 1.0 & 9.7 $\pm$ 0.2 \\
    0.10 & 0.05 & 3.8 $\pm$ 0.2 & 3.9 $\pm$ 0.1 & 0.3 $\pm$ 1.0 & 2.6 $\pm$ 0.5 & 3.1 $\pm$ 0.6 & 9.4 $\pm$ 1.0 \\
    \bottomrule
    \end{tabular}
  \end{subtable}
\end{figure*}

\begin{figure}[!htb]
  \centering
  \hspace*{-1.75cm}
  \includegraphics[width=1.4\textwidth]{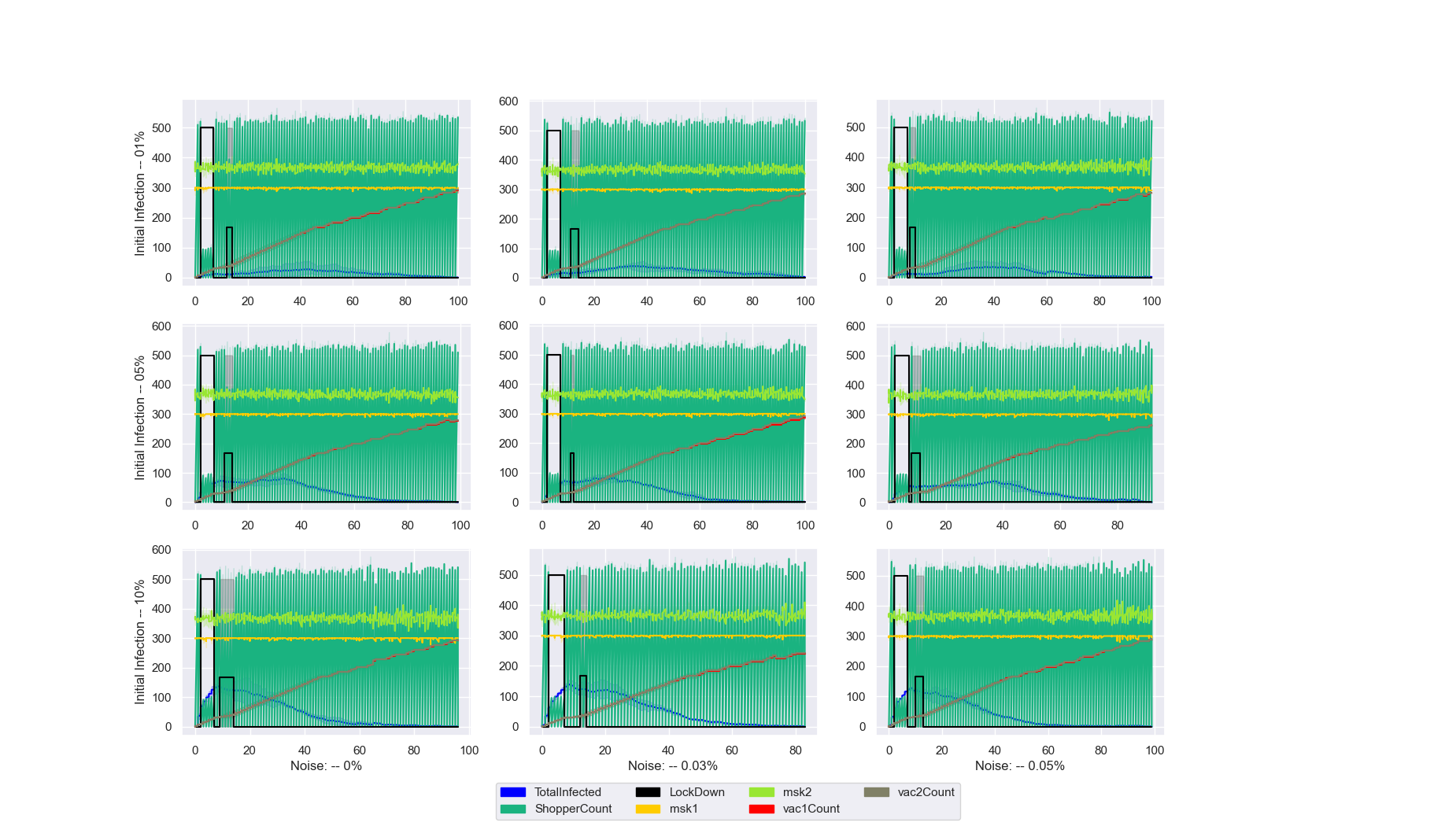}
   \caption{\textbf{High Vaccine Experiment} Detailed masking, vaccination, shopping and lockdown behavior}
  \label{fig:highvaccine_appendix}
\end{figure}

In this experiment, we wanted to check the response when low-efficiency vaccines became abundant (up to 100\% of the population). Uncertain action DDPG was the best-performing model that could effectively control both the infections and the economy ``(Figure \ref{fig:lowvaccine_epi}, Table \ref{tab:lowvaccine_actions})''. The results were similar to the basine and high-vaccine experiments ``(Figure \ref{fig:highvaccine_appendix})''. Once thing to note here is that the while both the baselines and the high vaccine scenarios maintained greater than 60\% vaccination rates irrespective of the initial infection rate. When higher levels of low-efficiency vaccinations were only available, the vaccination rates dropped to as low as 40\% of the population with an increase in the initial infection. 

In the best-performing models, across all experiments, despite starting with a lockdown in place, it was almost always removed either immediately after the first week or at the latest in two weeks. Repeat Lockdowns were imposed only in case of a high increase in the peak for a short period to flatten the peak. Similarly, the age groups 60-99 were most vaccinated (4 days per week), the age groups 18-59 were vaccinated for nearly two days initially, and the age 0-17 were vaccinated the least (less than 1 day per week). We observed that the rate of vaccination was constant across all the days. So a larger number of days translated to higher vaccination till about 40-50 days then the vaccination rate slightly dropped as more people were either vaccinated or recovered. While this is in line with the model where the elderly are highly susceptible, and the younger populations are relatively immune, our model differs from the widely practiced strategy to vaccinate the older populations first. Our model vaccinated the populations that had the most travel first. Despite not optimizing for death and total infection, the optimization could effectively control these factors. Across the different experiments, we saw that masking and the availability of high-efficiency masks had the most impact on controlling the epidemic. Though the vaccination did not decrease the peak's height, it made it narrower. There was not much difference between the low and high vaccine experiments in the control of the epidemic, but the individuals did not choose the lower quality vaccines resulting in lower vaccinations. Across the experiments, we also see that although the vaccination alone could not significantly impact when the initial infection load was high, they were very effective when started earlier in the pandemic. The actions and conclusions drawn from our results are in line with other studies on COVID-19 epidemics, which promote high vaccinations for the elderly (60-99) and then followed by the 18-59 group \cite{matrajt2021vaccine}.  Our model advocated for vaccinations of 18-59 first, followed by a longer vaccination of 60-99. In addition, imposing lockdowns also severely impacted the economy; sometimes, these were semi-permanent at best.

\subsection{Scenarios of Interest: Emergent Behavior}
The overarching goal of the individual decision-making model behavior to minimize the infections led to very interesting emergent behaviors such as panic shopping, masking related to infection, and masking correlated with shopping.

\begin{figure*}
\centering
\begin{multicols}{2}
    \includegraphics[width=0.99\linewidth]{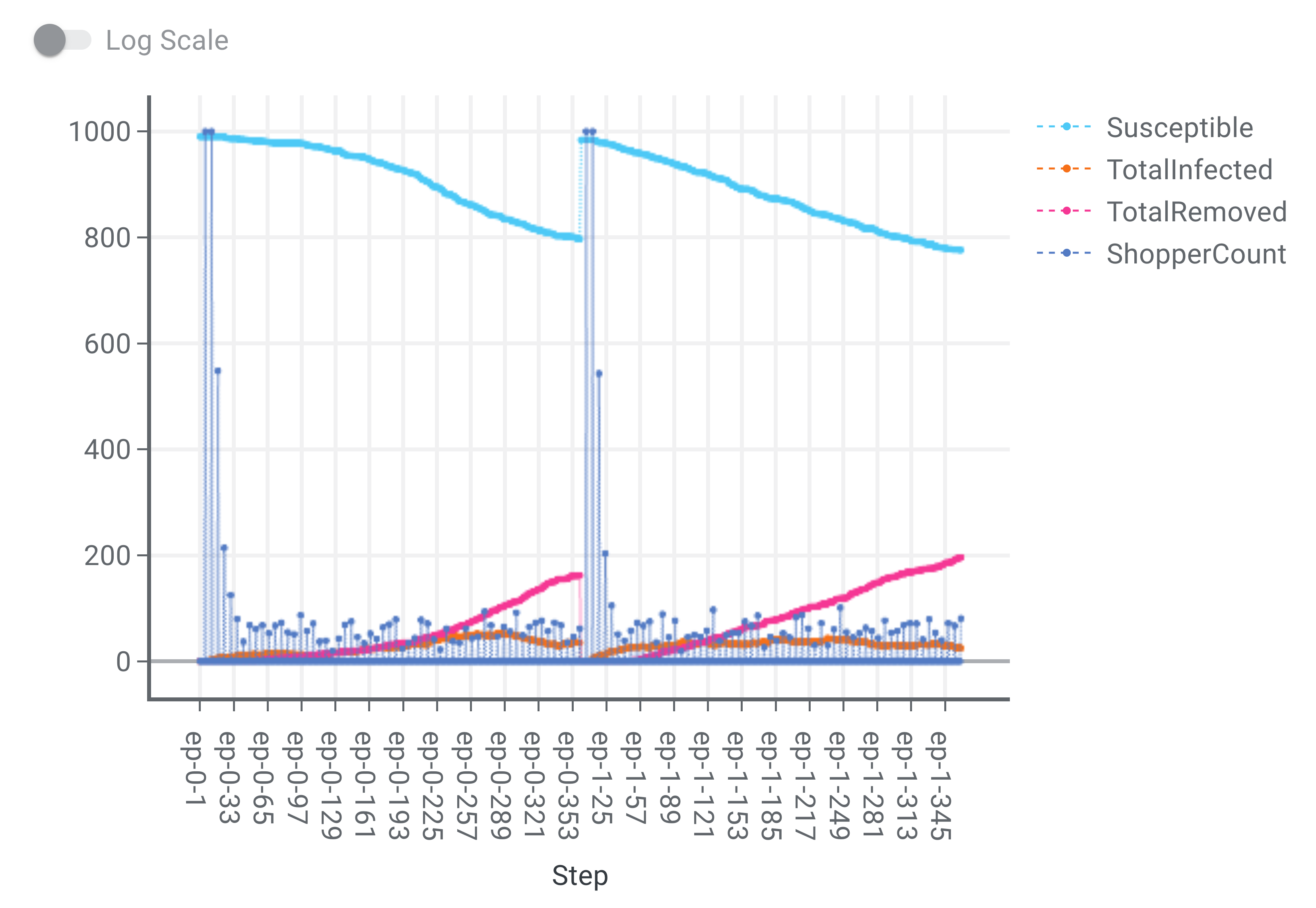}
    \subcaption{All shopping is at the beginning of the epidemic} \label{fig:panic_buying}
    \par 
    \includegraphics[width=0.99\linewidth]{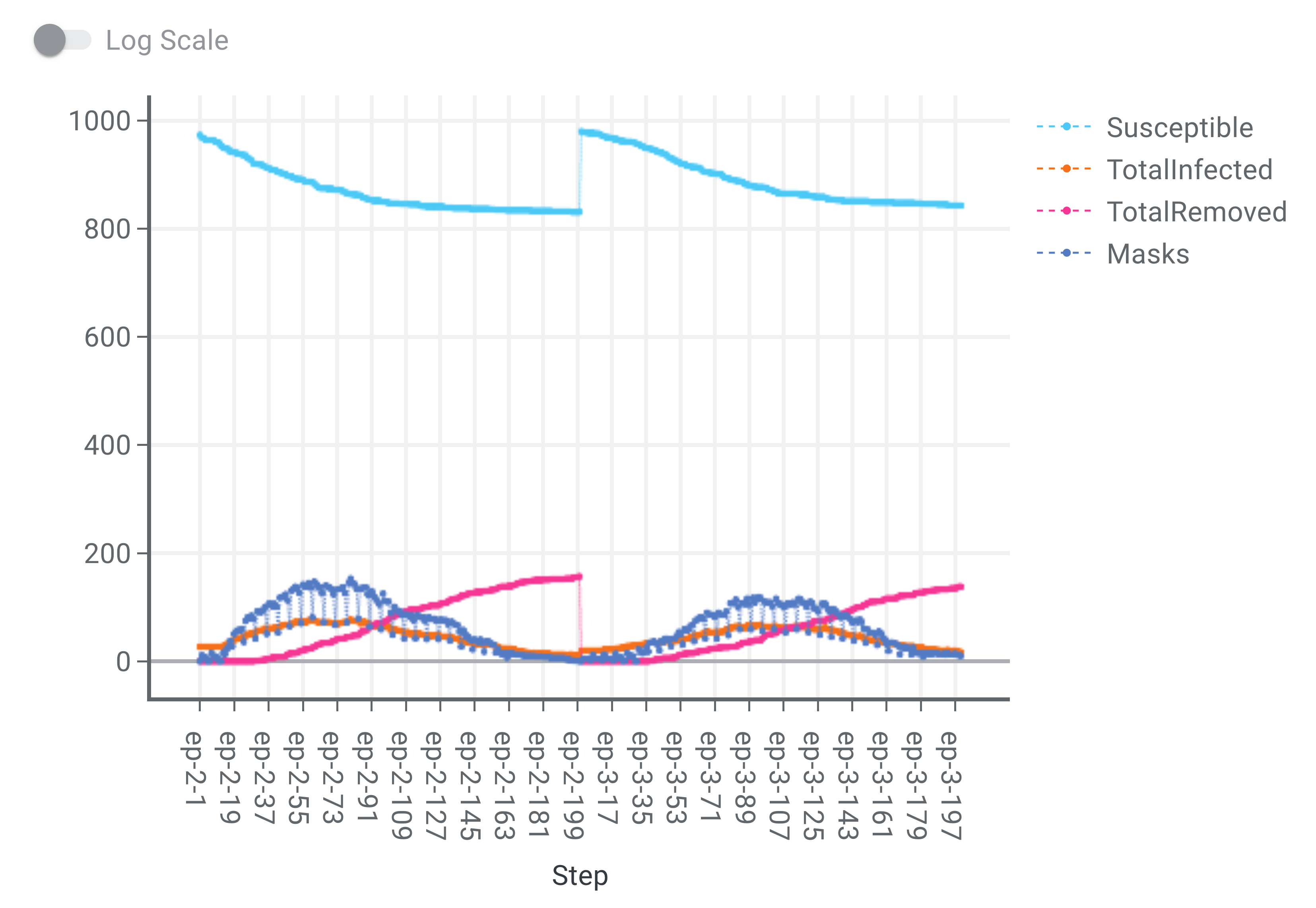}
    \subcaption{Mask decreases with slight reduction of infection} \label{fig:mask_fatigue} \par 
    \end{multicols}
    \subcaption{Correlation of masking with shopping counts}
    \includegraphics[width=0.99\linewidth]{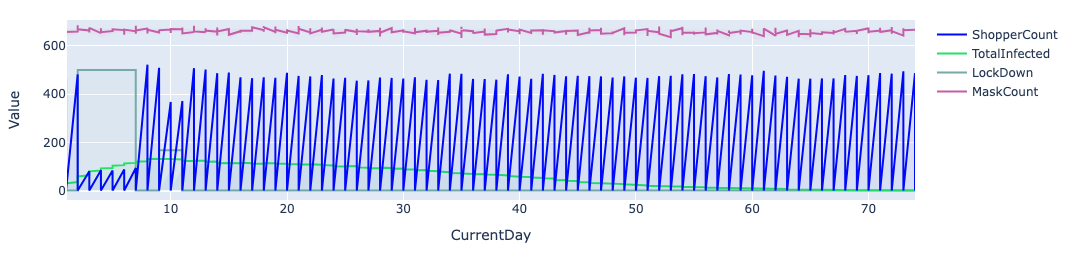} \label{fig:mask_shopping_corr} \par
\end{figure*}

\textbf{Panic shopping}
This behavior was observed more in the initial training stages when lockdowns were experimented with and not in any particular scenario. Across all scenarios, after training, the shoppers shopped when they needed supplies, and a lockdown correlated with lower shopping but not the absence of it. ``Figure \ref{fig:panic_buying}'', describes a panic shopping scenario where all shopping is at the beginning of the epidemic. We consistently saw people shop excessively despite negative rewards to discourage the event. this was because the shopping-related negative rewards were much lower than the disease-related ones.

\textbf{Infection dependent masking}
People had lower use of masks with lower infections ``(Figure \ref{fig:mask_fatigue})'', where many people stopped wearing masks as the infections decreased slightly. this was in line with some of the mask fatigue behavior seen in real life. This was again to reduce the negative rewards accrued from the continuous use of a mask.

\textbf{Correlation of masking with shopping counts}
Consistently across all the scenarios, there was a small peak in the mask count at the same time as a peak in the shopping counts ``(Figure \ref{fig:mask_shopping_corr})''. This is similar to the use of masks when going to new places.

\textbf{Lower preference for low-efficiency vaccines}
 Another interesting behavior we observed was that the people tried to choose masks when they were available and vaccinated themselves. However, when more low-efficiency vaccines were available, many people preferred to not vaccinate despite having similar outcomes on the pandemic. This was because in our model, the agents either got the vaccine they chose or they got no vaccines. When high-efficiency vaccines were chosen there were a large number of people who did not get vaccines (in the low-vaccine experiment). But despite that, people chose the better vaccine and it resulted in low vaccinations.

 With these examples, we could observe the ability of our hierarchical framework to exhibit realistic human behavior.

\section{Conclusion} \label{conclusion}
The discovery of a balanced multi-front public policy for epidemics is particularly challenging in a multi-objective setting with conflicting priorities. The decision-support tools for epidemic control typically derive policy without accounting for individual preferences and behavior and the uncertainty surrounding the decision-making process. They are also limited in the scale of the population, model complexity, and the number of choices for interventions (discrete vs. continuous).

This work illustrates public health and economy optimization in the presence of rational individual choices and uncertainty in state and action. We use and extend hierarchical reinforcement learning agents consisting of deep Q-networks and uncertainty-aware deep deterministic policy gradient variants (DDPG and TD3) for individual decisions and the policymaker's decision. We showcase five experimental settings on a minimalist model of a small community with different masks and vaccine availability. With those experiments, we can conclude that the use of hierarchical RL frameworks shows promise in the field of policy informatics and can effectively aid policymakers and computational epidemiologists by modeling realistic human behavior, such as panic shopping and lowered mask use with low infection. We demonstrated our results using a single model in a small community. In the near future, we will expand the model to include a larger scale of rational agents with more complex human and economic behavior. In addition, we do not employ explainability or techniques to explain each model's decision. In the near future, we aim to extend our work to account for explainability.

In conclusion, the current study significantly advances the field of agent-based epidemiological modeling and policy optimization, offering novel contributions that enhance the understanding of disease transmission dynamics and provide valuable decision support for policymakers. The integration of uncertainty-handling techniques and advanced metrics sets the stage for more reliable and adaptable policy recommendations in the realm of infectious disease management. As we extend our model to encompass larger communities and diverse landscapes, the implications of this research are poised to shape more effective and contextually relevant public health interventions in the future.

\section*{Declarations}

\begin{itemize}
\item Funding: There was no external funding to declare for this manuscript.
\item Competing interests: None of the authors has any financial affiliations that may be perceived to have biased the presentation. 
\item Ethics approval: Not Applicable
\item Consent to participate: Not applicable
\item Consent for publication: All the authors give their consent for publication
\item Availability of data and materials: We did not use any external datasets. The data for training  and testing of the models was generated from the epidemic simulation described in the paper.
\item {Authors' contributions: All the authors determined the problem statement after discussion with subject matter experts. This team then conceptualized the scope of the solution for this study and hence the subsequent publication. This was followed by the development of simulation models using the BharatSim framework by G.Deshkar and J. Kshirsagar. Following this, the RL modules were architected and developed for the individual agents and the policymakers by J. Venugopalan, and G. Deshkar. R. Gaur and J. Venugopalan conceptualized the evaluation of RL models which was then implemented by R. Gaur. All the authors came up with the study design for demonstrating the efficacy of the developed solution towards automated policy discovery, which was then implemented by the authors. All the authors contributed to the analysis of the study and the writing of the main manuscript. Finally, all the authors reviewed the manuscript.}
\end{itemize}
\endparano



% ...
\section{Appendix}
\subsubsection{Uncertain Action Algorithms}
\noindent
\begin{minipage}[t]{\textwidth}
\begin{algorithm} [H]
\caption{Uncertain Action DDPG Algorithm and Update}\label{alg:nrmdp_ddpg}
\footnotesize
\begin{algorithmic}[1]

\State \textbf{Input: } Actor update steps ($N$), uncertainty value $\alpha$ and discount factor $\gamma$

\State \textbf{Initialize: } Randomly initialize critic network $Q(s,a;\phi)$, actor $f(s;\theta)$ and adversary $\bar{f}(s;\bar{\theta})$

\State Initialize target networks $Q'(s,a;\phi^{-})$, $f'(s;\theta^{-})$ and $\bar{f}'(s;\bar{\theta}^{-})$ with weights $\phi^{-}$, $\theta^{-}$ and $\bar{\theta}^{-}$

\State Initialize replay buffer Replay

\For{episode in $0...M$}
 \State Reset to initial state $s_{0}$
 \For{$t = 1, T$}
  \State Sample action $a_{t} \sim \pi^{*}$; $a_{t} = (1 - \alpha)b + \alpha \bar{b}$
  \State $\tilde{a}_{t} = a_{t} +$ exploration noise
  \State Execute $\tilde{a}_{t}$ and observe reward $r_{t}$ and new state $s_{t+1}$
  \State Store transition ($s_{t}, \tilde{a}_{t}, r_{t}, s_{t+1}$) in Replay
  \State Sample a random mini-batch of transitions from replay
  \State Set $ y_{i} = r + \gamma[Q'(s',(1 - \alpha)f'(s;\theta^{-}) + \alpha \bar{f}'(s;\bar{\theta}^{-})]$
  \State Update critic with critic loss $L = \frac{1}{N}\sum_{i}[y_{i} - Q(s_{i},a_{i})]^2$
  \If{episode = $N$}
    \State Update actor $\theta \gets \nabla_{\theta} Q(s,(1 -\alpha)f(s;\theta) + \alpha\bar{f}(s;\bar{\theta}))$
    \State Update adversary 
    \State$\bar{\theta} \gets \nabla_{\bar{\theta}} Q(s,(1 -\alpha)f(s;\theta) + \alpha\bar{f}(s;\bar{\theta}))$
    \State Update the target networks
    \State $\phi^{-} \gets \tau\phi + (1-\tau)\phi^{-}$
    \State $\theta^{-} \gets \tau\theta + (1-\tau)\theta^{-}$
    \State $\bar{\theta}^{-} \gets \tau\bar{\theta} + (1-\tau)\bar{\theta}^{-}$
  \EndIf
 \EndFor
\EndFor

\end{algorithmic}
\end{algorithm}
\end{minipage}

\subsubsection{Uncertain State and Action Algorithms}
\begin{minipage}[t]{\textwidth}
\begin{algorithm}[H]
\caption{Uncertain State and Action DDPG Algorithm and Update}\label{alg:nrmdp_ddpg_bayes}
\footnotesize
\begin{algorithmic}[1]

\State \textbf{Input: } Actor update steps ($N$), uncertainty value $\alpha$, discount factor $\gamma$ and samples $S$

\State \textbf{Initialize: } Randomly initialize critic network $Q(s,a;\phi) \sim N(0,1)$, actor $f(s;\theta)$ and adversary $\bar{f}(s;\bar{\theta})$

\State Initialize target networks $Q'(s,a;\phi^{-})$, $f'(s;\theta^{-})$ and $\bar{f}'(s;\bar{\theta}^{-})$ with weights $\phi^{-}$, $\theta^{-}$ and $\bar{\theta}^{-}$

\State Initialize replay buffer Replay

\For{episode in $0...M$}
 \State Reset to initial state $s_{0}$
 \For{$t = 1, T$}
  \State Sample action $a_{t} \sim \pi^{*}$; $a_{t} = (1 - \alpha)b + \alpha \bar{b}$
  \State $\tilde{a}_{t} = a_{t} +$ exploration noise
  \State Execute $\tilde{a}_{t}$ and observe reward $r_{t}$ and new state $s_{t+1}$
  \State Store transition ($s_{t}, \tilde{a}_{t}, r_{t}, s_{t+1}$) in Replay
  \State Sample a random mini-batch of transitions from replay
  \State Set $ y_{i} = r + \gamma[Q'(s',(1 - \alpha)f'(s;\theta^{-}) + \alpha \bar{f}'(s;\bar{\theta}^{-})]$
  \State where we sample  weights $\phi^{-}$ $S$ times and compute mean
  \State Update critic with critic loss $L = \frac{1}{N}\sum_{i}[y_{i} - Q(s_{i},a_{i})]^2$
  State where we sample  weights $\phi$ $S$ times and compute mean
  \If{episode = $N$}
    \State Update actor $\theta \gets \nabla_{\theta} Q(s,(1 -\alpha)f(s;\theta) + \alpha\bar{f}(s;\bar{\theta}))$
    \State Update adversary 
    \State$\bar{\theta} \gets \nabla_{\bar{\theta}} Q(s,(1 -\alpha)f(s;\theta) + \alpha\bar{f}(s;\bar{\theta}))$
    \State Update the target networks
    \State $\phi^{-} \gets \tau\phi + (1-\tau)\phi^{-}$
    \State $\theta^{-} \gets \tau\theta + (1-\tau)\theta^{-}$
    \State $\bar{\theta}^{-} \gets \tau\bar{\theta} + (1-\tau)\bar{\theta}^{-}$
  \EndIf
 \EndFor
\EndFor

\end{algorithmic}
\end{algorithm}
\end{minipage}

\noindent
\begin{minipage}[t]{\textwidth}
\begin{algorithm}[H]
\caption{Uncertain State and Action TD3 Algorithm and Update}\label{alg:nrmdp_td3_bayes}
\footnotesize
\begin{algorithmic}[1]

\State \textbf{Input: } Actor update steps ($N$), uncertainty value $\alpha$, discount factor $\gamma$ and samples $S$

\State \textbf{Initialize: } Randomly initialize critic network $Q_{1,2}(s,a;\phi) \sim N(0, 1)$, actor $f(s;\theta)$ and adversary $\bar{f}(s;\bar{\theta})$

\State Initialize target networks $Q'_{1,2}(s,a;\phi^{-})$, $f'(s;\theta^{-})$ and $\bar{f}'(s;\bar{\theta}^{-})$ with weights $\phi^{-}$, $\theta^{-}$ and $\bar{\theta}^{-}$

\State Initialize replay buffer Replay

\For{episode in $0...M$}
 \State Reset to initial state $s_{0}$
 \For{$t = 1, T$}
  \State Sample action $a_{t} \sim \pi^{*}$; $a_{t} = (1 - \alpha)b + \alpha \bar{b}$
  \State $\tilde{a}_{t} = a_{t} +$ exploration noise
  \State Execute $\tilde{a}_{t}$ and observe reward $r_{t}$ and new state $s_{t+1}$
  \State Store transition ($s_{t}, \tilde{a}_{t}, r_{t}, s_{t+1}$) in Replay
  \State Sample a random mini-batch of transitions from replay
  \State Set $ y_{i} = r + \gamma[min(Q'(s',(1 - \alpha)f'(s;\theta^{-}) + \alpha \bar{f}'(s;\bar{\theta}^{-}))]$
  \State where we sample  weights $\phi^{-}$ $S$ times and compute mean
  \State Update critic with critic loss 
  \State $L = \frac{1}{N}\sum_{i}([y_{i} - Q_{1}(s_{i},a_{i})]^2 + [y_{i} - Q_{2}(s_{i},a_{i})]^2)$
  \State where we sample  weights $\phi$ $S$ times and compute mean
  \If{episode = $N$}
    \State Update actor $\theta \gets \nabla_{\theta} Q_{1}(s,(1 -\alpha)f(s;\theta) + \alpha\bar{f}(s;\bar{\theta}))$
    \State Update adversary 
    \State$\bar{\theta} \gets \nabla_{\bar{\theta}} Q_1{1}(s,(1 -\alpha)f(s;\theta) + \alpha\bar{f}(s;\bar{\theta}))$
    \State Update the target networks
    \State $\phi^{-} \gets \tau\phi + (1-\tau)\phi^{-}$
    \State $\theta^{-} \gets \tau\theta + (1-\tau)\theta^{-}$
    \State $\bar{\theta}^{-} \gets \tau\bar{\theta} + (1-\tau)\bar{\theta}^{-}$
  \EndIf
 \EndFor
\EndFor

\end{algorithmic}
\end{algorithm}
\end{minipage}

\subsubsection{State and Action Binning Algorithms}

\noindent
\begin{minipage}[t]{\textwidth}
\begin{algorithm}[H]
\caption{State Binning Algorithm}\label{alg:state_binning}
\footnotesize
\begin{algorithmic}[1]

\State \textbf{Input: } Path of the CSV file.
\vspace{5mm}
\For{policy in $0...P$}
    \State \textbf{Step 1: } Extract rows $(R)$ of output columns for a policy from \indent \indent  CSV.
    \vspace{5mm}
    \State \textbf{Step 2: } Compute state $(S)$ using: 
    \State \hspace{1cm}$State [S_{1}, S_{2}, S_{3}] = \frac{Mean(R) + Max(R)}{N}$
    \State \hspace{1cm}Number of Agents (N) = 1000 in this study.
    \vspace{5mm}
    \State \textbf{Step 3: } Now for computed state $(S)$ perform binning, \indent \indent and store bin value for each state in $S$ in a list $(S_{b})$.
    \vspace{5mm}
    \State \textbf{Step 4: } Now, generate $N$ possible combinations of values
    \indent \indent corresponding to $S_{b}$.
    \vspace{5mm}
    \State \textbf{Step 5: } Return index of combination $c$ which is equal to $S_{b}$ \indent \indent from $N$ possible combinations as State index $(S_{i})$.
\EndFor
\end{algorithmic}
\end{algorithm}
\end{minipage}

\noindent
\begin{minipage}[t]{\textwidth}
\begin{algorithm}[H]
\caption{Action Binning Algorithm}\label{alg:action_binning}
\footnotesize
\begin{algorithmic}[1]

\State \textbf{Input: } Path of output folder.
\vspace{5mm}
\For{policy in $0...P$}
    \State \textbf{Step 1: } Extract actions from policy and store them in a list $(A)$.
    \vspace{5mm}
    \State \textbf{Step 2: } Perform binning for each action in $A$ and store bin value \indent \indent for each action in a list $(A_{b})$.
    \vspace{5mm}
    \State \textbf{Step 3: } Now, generate $N$ possible combinations of values
    \indent \indent corresponding to $A_{b}$.
    \vspace{5mm}
    \State \textbf{Step 4: } Return index of combination $c$ which is equal to $A_{b}$ \indent \indent from $N$ possible combinations as Action index $(A_{i})$.
\EndFor
\end{algorithmic}
\end{algorithm}
\end{minipage}

\begin{figure}
  \centering
  \includegraphics[width=0.55\textwidth]{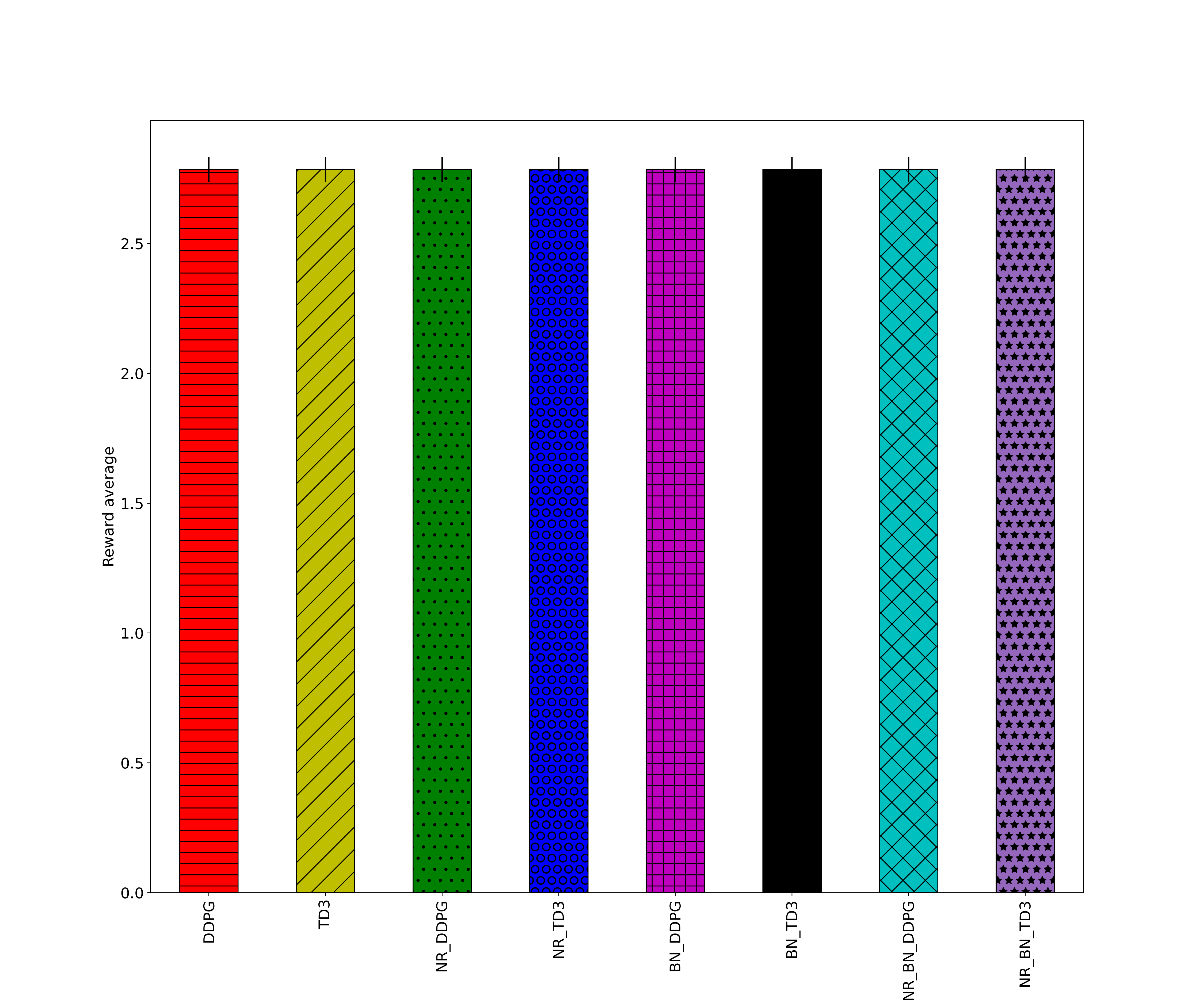}
  \caption{Overall average reward shows all algorithms to have similar rewards.}
  \label{fig:overall_avg_reward}
\end{figure}

\begin{table*}
\tiny
\caption{Reward Analysis}
\label{tab:reward_analysis}

\begin{tabular}{|c|c|c|c|c|c|c|c|c|c|c|c|}
\toprule
Outlier &        Inf &       Hosp &  Stock &   $L_{Start}$ &     $L_{Dur}$ &  $V^{0-17}_{Start}$ &    $V^{0-17}_{Dur}$ &  $V^{18-59}_{Start}$ &    $V^{18-59}_{Dur}$ &  $V^{60-99}_{Start}$ &    $V^{60-99}_{Dur}$ \\
\midrule
  Lower & 0.0-0.33 & 0.0-0.33 &  0.67-1.0 & 3.5-7.0 & 0.0-3.5 & 3.5-7.0 & 0.0-3.5 & 3.5-7.0 & 0.0-3.5 & 3.5-7.0 & 0.0-3.5 \\
  Lower & 0.0-0.33 & 0.0-0.33 &  0.0-0.33 & 3.5-7.0 & 0.0-3.5 & 3.5-7.0 & 0.0-3.5 & 3.5-7.0 & 0.0-3.5 & 3.5-7.0 & 3.5-7.0 \\
  Lower & 0.0-0.33 & 0.0-0.33 &  0.67-1.0 & 3.5-7.0 & 0.0-3.5 & 3.5-7.0 & 0.0-3.5 & 3.5-7.0 & 0.0-3.5 & 3.5-7.0 & 3.5-7.0 \\
  Upper & 0.0-0.33 & 0.0-0.33 &  0.0-0.33 & 3.5-7.0 & 0.0-3.5 & 3.5-7.0 & 0.0-3.5 & 3.5-7.0 & 0.0-3.5 & 3.5-7.0 & 3.5-7.0 \\
  Upper & 0.0-0.33 & 0.0-0.33 &  0.0-0.33 & 0.0-3.5 & 0.0-3.5 & 0.0-3.5 & 0.0-3.5 & 0.0-3.5 & 0.0-3.5 & 0.0-3.5 & 0.0-3.5 \\
  Upper & 0.0-0.33 & 0.0-0.33 &  0.67-1.0 & 0.0-3.5 & 0.0-3.5 & 0.0-3.5 & 0.0-3.5 & 0.0-3.5 & 0.0-3.5 & 0.0-3.5 & 0.0-3.5 \\
  Upper & 0.0-0.33 & 0.0-0.33 &  0.0-0.33 & 3.5-7.0 & 0.0-3.5 & 3.5-7.0 & 0.0-3.5 & 3.5-7.0 & 0.0-3.5 & 3.5-7.0 & 0.0-3.5 \\
  Upper & 0.0-0.33 & 0.0-0.33 &  0.67-1.0 & 3.5-7.0 & 0.0-3.5 & 3.5-7.0 & 0.0-3.5 & 3.5-7.0 & 0.0-3.5 & 3.5-7.0 & 0.0-3.5 \\
  Upper & 0.0-0.33 & 0.0-0.33 & 0.33-0.67 & 0.0-3.5 & 0.0-3.5 & 0.0-3.5 & 0.0-3.5 & 0.0-3.5 & 0.0-3.5 & 0.0-3.5 & 0.0-3.5 \\
\bottomrule
\multicolumn{12}{l}{\textbf{Note:} $Inf$: InfectedMild, $Hosp$: Hospitalized, $L$: Lockdown, $Dur$: Duration; $Stock$ : House Stock}
\end{tabular}
\end{table*}

\begin{figure}[!htb]
  \centering
  \hspace*{-1.75cm}
  \includegraphics[width=1.4\textwidth]{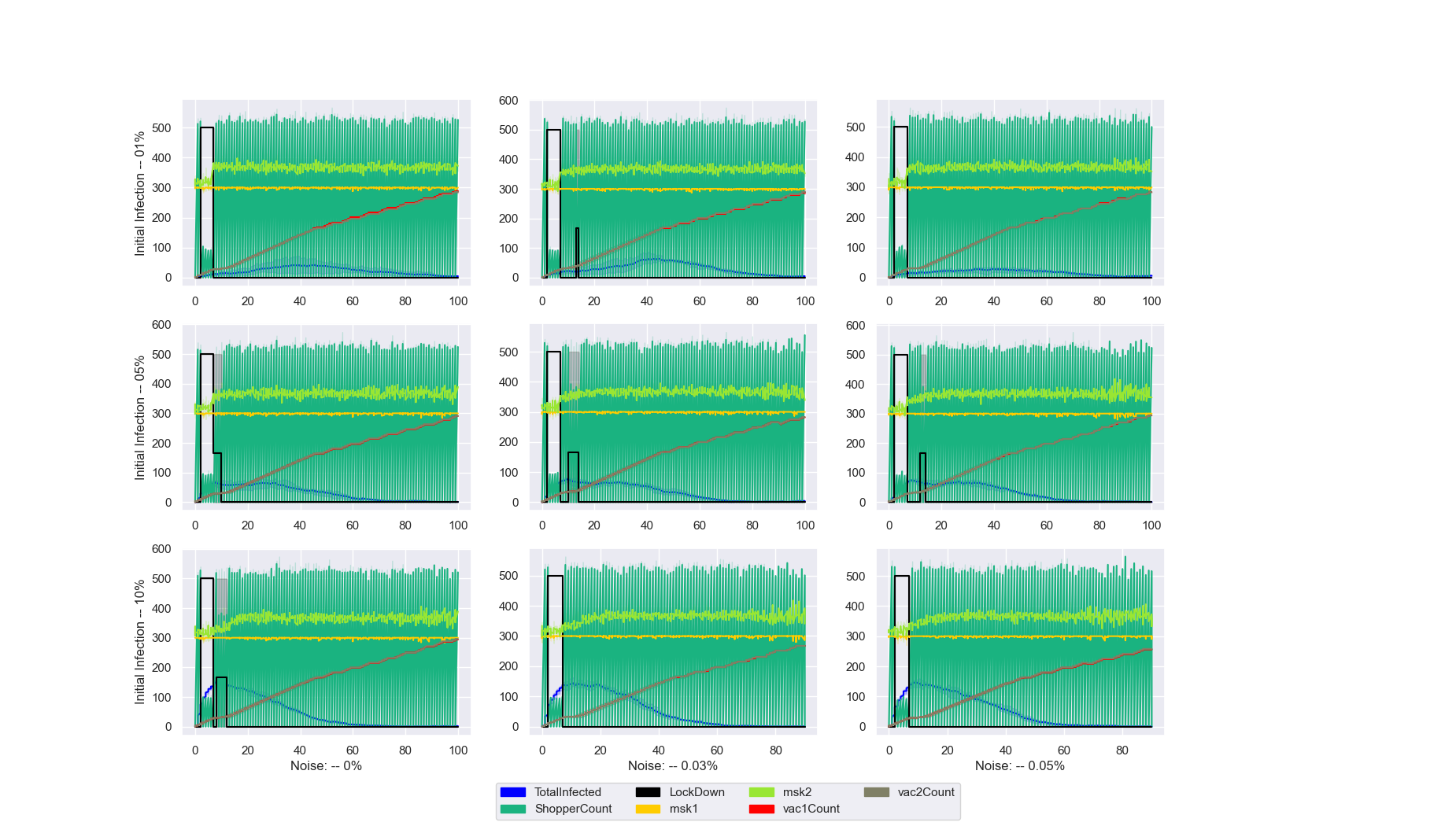}
   \caption{\textbf{Low Mask Experiment} Detailed masking, vaccination, shopping and lockdown behavior}
  \label{fig:lowmask_appendix}
\end{figure}

\begin{figure}[!htb]
  \centering
  \hspace*{-1.75cm}
  \includegraphics[width=1.4\textwidth]{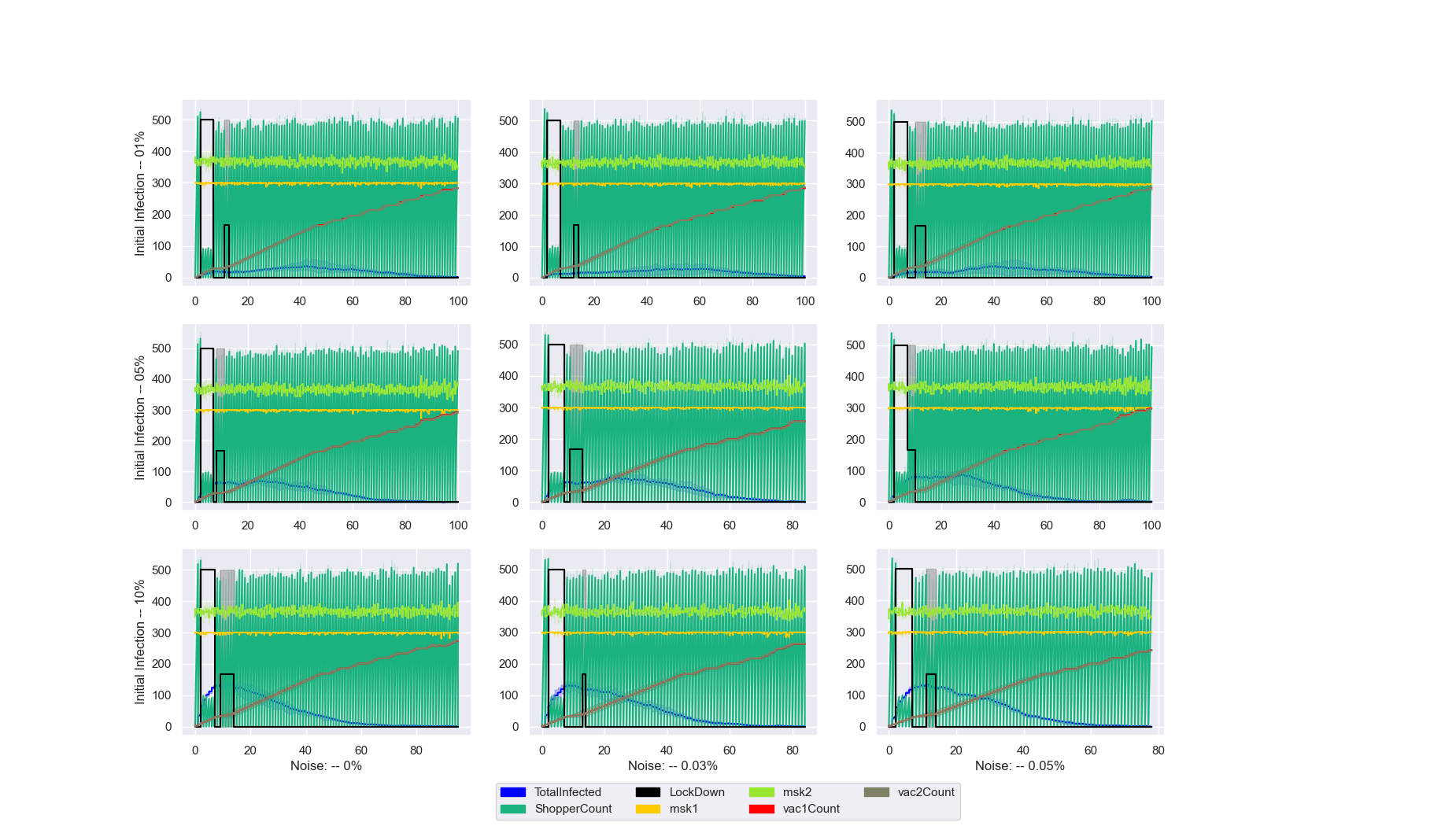}
   \caption{\textbf{Low Vaccine Experiment} Detailed masking, vaccination, shopping and lockdown behavior}
  \label{fig:lowvaccine_appendix}
\end{figure}




 
\bibliographystyle{jasss}
\bibliography{bibliography} 


\end{document}